\documentclass[11pt]{article}

\usepackage{graphicx}
\usepackage{amsmath,amsfonts,amssymb,bm}
\usepackage{mathptmx}
\usepackage{newtxtext}
\usepackage{newtxmath}
\usepackage{epstopdf}
\usepackage[scaled=.97]{helvet}
\usepackage{fancyhdr}
\usepackage{natbib}
\usepackage{url}
\usepackage{xcolor}
\usepackage{indentfirst}
\usepackage{multicol}
\usepackage{ifthen}
\usepackage{rotating}
\usepackage{appendix}
\usepackage{setspace}
\usepackage{lineno}
\usepackage{authblk}
\usepackage{hyperref}
\usepackage{tabularx}

\title{\bfseries A Hybrid LSTM–Vision Transformer Architecture for Predicting HRRR Forecast Errors}

\author[1]{D. Aaron Evans\thanks{Corresponding author: \href{mailto:aaevans@albany.edu}{aaevans@albany.edu}}}
\author[3]{Jay C. Rothenberger}
\author[1]{Kara J. Sulia}
\author[2]{Nick P. Bassill}
\author[1]{Chris D. Thorncroft}

\affil[1]{Atmospheric Sciences Research Center, University at Albany, SUNY, Albany, NY, USA}
\affil[2]{State Weather Risk Communication Center, University at Albany, SUNY, Albany, NY, USA}
\affil[3]{University of Oklahoma, Norman, OK, USA}

\date{22 June 2026}

\begin{document}
\maketitle

%% ABSTRACT %%
\begin{abstract}
Forecast errors in high-resolution numerical weather prediction (NWP) systems are often linked to unresolved planetary boundary layer (PBL) processes, convection, terrain-induced circulations, and other vertically structured atmospheric phenomena. Previous work demonstrated that Long Short-Term Memory (LSTM) networks can successfully predict forecast errors in the High-Resolution Rapid Refresh (HRRR) model using mesonet observations, but we believe performance degradation is linked to periods of complex vertical atmospheric evolution. To address this limitation, we develop a hybrid LSTM–Vision Transformer (LSTM–ViT) framework that combines temporal sequence learning from surface observations with atmospheric profiles from the New York State Mesonet profiler network. The LSTM–ViT framework is trained to predict HRRR hourly precipitation, 10-m wind speed, and 2-m temperature forecast errors at individual mesonet stations. Across all three predictors, incorporation of profiler-derived atmospheric structure improves forecast-error prediction skill relative to the baseline LSTM architecture, with the largest gains occurring at shorter forecast lead times and during periods of enhanced PBL activity. Improvements are particularly pronounced for precipitation forecast error, where the LSTM-ViT framework achieves approximately a twofold increase in predictive skill relative to the baseline LSTM while better capturing convectively driven error evolution and reducing degradation associated with PBL processes. These results demonstrate that combining temporal sequence learning with vertically informed attention mechanisms provides a physically meaningful pathway for improving forecast-error prediction in operational NWP systems. Our research offers forecasters enhanced guidance regarding model bias and forecast confidence.\footnote{This manuscript is a preprint and has been submitted for peer review to the \textit{Artificial Intelligence for the Earth Systems} journal. The content is subject to change based on the outcome of the peer-review process and should not be considered final or definitive. Copyright in this Work may be transferred without further notice.}
\end{abstract}

\section{Introduction}

Numerical weather prediction (NWP) systems exhibit limited resolution and systematic forecast errors that are often difficult to represent explicitly within traditional modeling frameworks, motivating the development of machine-learning (ML) methods to predict and correct these errors using historical forecasts and observations. Early ML applications demonstrated value in post-processing and decision-support contexts, including real-time guidance for high-impact weather in the Great Plains \citep{McGovern2017} and probabilistic prediction of hail occurrence and size \citep{Gagne2017}. Building on these successes, ML-driven forecast products such as GraphCast \citep{Lam2022} and ECMWF’s Artificial Intelligence Forecasting System (AIFS; \citealp{Lang2024}) now provide full-field, medium-resolution global forecasts, marking a significant shift toward ML-augmented operational weather prediction.

Persistent errors in high-resolution NWP systems continue to motivate higher fidelity and more expedient post-processing techniques that improve forecast quality without modifying the underlying model. To address this challenge, \citet{Evans2025} developed a Long Short-Term Memory (LSTM) framework that predicts forecast error directly from NWP forecasts and surface observations. The LSTM models improve baseline High-Resolution Rapid Refresh (HRRR) forecasts by 15--75\%, depending on target variable, demonstrating that ML can learn systematic forecast-error characteristics from historical data. Because the framework operates as a post-processing step, forecasters can apply it to existing NWP products without additional numerical integrations, providing a practical means of assessing forecast uncertainty and correcting forecast biases at the point of use. The widespread availability of dense surface observations from more than 30 statewide mesonet networks across the United States \citep{Mahmood2017, MADISMesonetProviders} also makes the approach readily transferable across diverse forecasting environments.

In \citet{Evans2025}, we evaluate LSTM performance across the New York State Mesonet (NYSM) and Oklahoma State Mesonet using mean absolute error (MAE) and mean error metrics computed between predicted and observed NWP forecast errors, while examining skill across geographic regions, diurnal cycles, seasonal cycles, and relative improvements over HRRR forecasts. These analyses link systematic performance degradation along mesoscale boundaries associated with topographic influences and latent atmospheric processes. The strongest degradation is correlated with periods of active or complex planetary boundary layer (PBL) evolution, suggesting that unresolved boundary-layer processes represent an important limitation of the surface-based LSTM framework. The present work addresses this limitation by incorporating vertically resolved PBL observations from microwave radiometers through a Vision Transformer (ViT) encoder, enabling the model to learn atmospheric structures above the surface that may contribute to forecast error.

Previous studies have demonstrated the value of combining recurrent and attention-based representations for complex forecasting problems. For example, \citet{Zhang2025Hybrid} found that hybrid LSTM–Transformer architectures consistently outperformed standalone LSTM and Transformer models for electricity load forecasting, while \citet{tang2023} showed that SwinLSTM outperformed ConvLSTM on spatiotemporal prediction tasks. Together, these studies demonstrate that attention mechanisms can complement recurrent sequence modeling by improving representations of complex spatial and temporal dependencies. However, both frameworks operated on a single data modality. In contrast, the present work combines surface observations, NWP forecast fields, and vertically resolved microwave-radiometer profiles within a unified architecture to improve forecast-error prediction in a high-resolution NWP environment.

This study evaluates a LSTM–Vision Transformer (LSTM-ViT) framework for predicting accumulated precipitation, 10-m wind magnitude, and 2-m temperature forecast errors within HRRR forecasts. We hypothesize that vertically resolved PBL observations contain information relevant to forecast-error evolution that surface observations alone cannot capture. Consequently, incorporating vertical profile information through a ViT encoder should improve forecast-error prediction skill relative to a surface-based LSTM architecture, particularly during periods of complex boundary-layer evolution and mesoscale forcing.

\section{Data}
\subsection{NYSM Standard Network}

Operational since 2018, the NYSM comprises 127 weather stations across New York State (NYS), with an average spacing of 27 kilometers \citep[hereafter B20]{Brotzge2020}. Note: Lake Placid Station is excluded, as it was installed in May 2024, which is outside of the training period. The NYSM is recognized for its rigorous quality control, precise instrument calibration, and strict site placement requirements (B20). NYSM data undergoes both automated and manual quality assurance processes in real time, as well as on a daily, weekly, monthly, and annual basis (B20). Each observation is automatically assigned a quality flag: good, suspect, warning, or failure. As such, the data used to train the ML models herein excludes data flagged with warning or failure. 
    
As established in \citet[hereafter E25]{Evans2025}, the NYSM observations are aligned with the temporal scale of the NWP model forecast. To align the temporal scale of the NYSM observations, which are recorded every five minutes, with that of an NWP model forecast, the observations taken at the top of each hour are used as the true observed atmospheric conditions during training. There are two exceptions to this: total precipitation is accumulated over the hour, and wind speed is averaged over the hour. There are 16 meteorological variables from the NYSM standard network that are used as features in training the LSTM models. These features are listed in Table \ref{tab:combined_features}.

\begin{table*}[ht]
\centering
\resizebox{\textwidth}{!}{%
\begin{tabular}{|l|l|l|}
\hline
\textbf{HRRR Model Features} & \textbf{New York State Mesonet Features} & \textbf{Profiler Features} \\
\hline\hline
-- & Latitude & --  \\
-- & Longitude & -- \\
-- & Elevation & -- \\
2-Meter Temperature & 2-Meter Temperature & Temperature \\
2-Meter Specific Humidity & 9-Meter Temperature &   Infrared Temperature \\
2-Meter Dew Point & 2-Meter Dew Point & Dew Point \\
2-Meter Relative Humidity & 2-Meter Relative Humidity & Relative Humidity \\
Downward SW Radiation & Solar Radiation & Liquid Content \\
Downward LW Radiation & Atmospheric Pressure & Pressure Level \\
Mean Sea-Level Pressure & Mean Sea-Level Pressure & Integrated Vapor \\
Total Wind Speed & Mean 10-Meter Sonic Anemometer Wind Speed & Integrated Liquid \\
10-Meter Wind U Component & 10-Meter Sonic Anemometer Wind Speed &  Vapor Density \\
10-Meter Wind V Component & Max 10-Meter Sonic Anemometer Wind Speed & -- \\
10-Meter Wind Direction & 10-Meter Wind Direction & -- \\
Total Hourly Precipitation & Total Hourly Precipitation & Rain Flag \\
Accumulated Snow & Snow Depth & -- \\
CAPE & -- & -- \\
Total Cloud Cover & -- &  Cloud Base \\
500-hPa Geopotential Height & -- & -- \\
\hline
\end{tabular}}
\caption{Combined list of HRRR, NYSM-Standard, and NYSM-Profiler independent variables used as features in training the Hybrid LSTM-ViT models.}
\label{tab:combined_features}
\end{table*}

\subsection{NYSM Profiler Network}

\begin{figure}
    \centering
    \includegraphics[width=33pc]{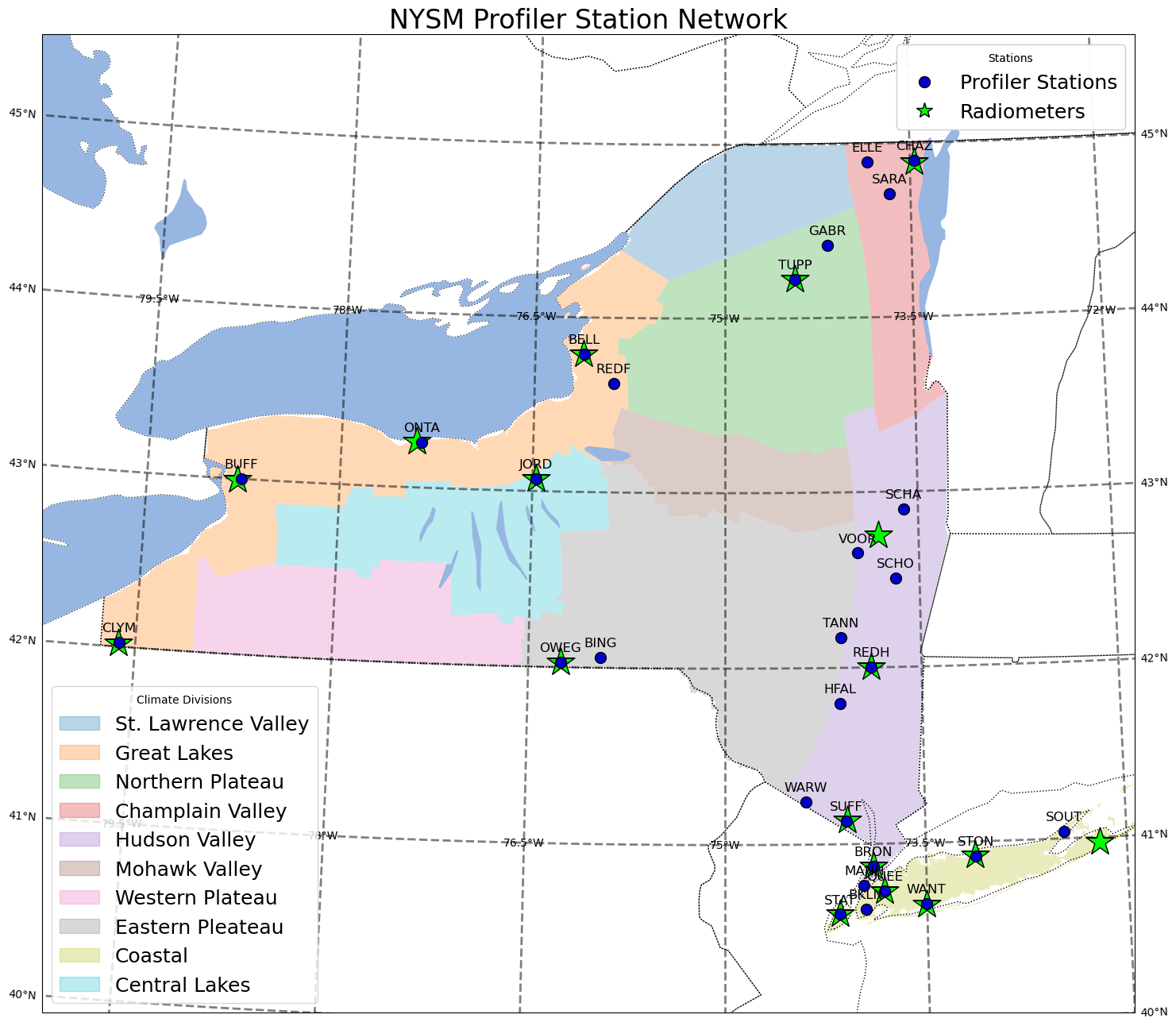}
    \caption{New York State overlaid with panels corresponding to NCEI climate divisions \citep{NCEI2015}. Green stars represent the locations of the microwave radiometers, and blue scatter points represent the locations of the NYSM stations eligible for use across the NYSM standard network.}\label{fig:nysm_profiler_network}
\end{figure}

The NYSM Profiler Network consists of 17 stations across the standard network \citep[hereafter S22; see Fig.~\ref{fig:nysm_profiler_network}]{Shrestha22}. For this study, we use data from the passive Radiometrics MP-3000 series microwave radiometers (MWR). The MWR operates across two spectral bands, the 21 K band (22–30 GHz) and the 14 V band (51–59 GHz), to measure brightness temperatures associated with the water vapor and oxygen absorption bands (S22). These brightness temperatures are then used to retrieve vertical profiles of atmospheric temperature, relative humidity, water vapor density, and liquid water content (S22). Retrievals are performed using a neural network in conjunction with a radiative transfer algorithm (S22). For the purposes of this work, vertical profiles extend from the surface to 5 km, a height sufficient to capture the PBL and mixing layer over NYS \citep{Shrestha2021, Molod2019}. The vertical resolution of the MWR is 50 m below 500 m, 100 m between 500 m and 2000 m, and 250 m above 2000 m (S22). The temporal resolution of the retrieved profiles is 10 minutes (S22).

Despite the utility of the MWR data, the retrieval process is inherently ill-posed and subject to multiple sources of error. Biases in the measured brightness temperatures can arise from uncertainties in the gas absorption model, calibration errors (such as liquid nitrogen reference loads), and limitations in the neural network algorithm (S22). Across NYSM sites, the MWR exhibits a consistent cold bias, with positive mean bias errors (MBE) ranging between 2.7°C and 3.3°C, values that are statistically significant (p $\leq$ 0.05) (S22). A key source of this bias is the instrument’s failure to detect elevated temperature inversion layers, which leads to a sharp increase in cold bias above the inversion (S22).

Water vapor density retrievals tend to be more accurate at altitudes above the PBL (S22). This likely reflects the inherently higher moisture variability near the surface and the fact that water vapor content decreases substantially with height (S22).

The MWR is also equipped with a precipitation sensor that detects whether precipitation is falling over the radiometer’s radome using a binary status flag (S22). Additionally, an infrared radiation thermometer measures the cloud base temperature (S22). Cloud base height is then defined as the lowest altitude where the retrieved temperature profile matches the cloud base temperature (S22).

Days with precipitation exhibit the lowest temperature retrieval errors across all error metrics (MBE, MAE, and RMSE), while clear-sky days exhibit the largest temperature errors (S22). Temperature errors tend to peak on clear-sky days, whereas water vapor density errors are highest during cloudy and precipitating conditions (S22). 

The MWR observations are structured to resemble image-like arrays, enabling compatibility with a Vision Transformer (ViT) encoder. Rather than using the conventional image format of Width $\times$ Height $\times$ Channels (e.g., red, green, blue), for each site, the MWR data are organized as Timestamp $\times$ Height $\times$ Variable. In this case, the “channels” correspond to atmospheric variables, as listed in Table \ref{tab:combined_features}. For each prediction, the model uses MWR observations from the most recent complete hour, which comprises 6 time steps, collected by the nearest profiler station.

\subsection{High-Resolution Rapid Refresh Forecast System}

HRRR forecast system, developed by the National Oceanic and Atmospheric Administration (NOAA) in 2014 \citep{Dowell2022}, employs a cloud-resolving, convection-allowing implementation of the Advanced Research version of the Weather Research and Forecasting (WRF-ARW) model as its dynamical core \citep{NCEP_HRRR}. HRRR was specifically designed to support short-range forecasting of rapidly evolving weather phenomena, with particular emphasis on convection and precipitating systems that are critical for operational situational awareness \citep{Dowell2022}. The model operates on a 3-km Lambert Conformal grid spanning the continental United States \citep{NCEP_HRRR} and is initialized hourly, providing forecasts out to 18 hours. Although HRRR also produces extended forecasts to 48 hours for the 00, 06, 12, and 18 UTC initialization cycles \citep{Dowell2022}, this study focuses on the first 18 forecast hours to maintain a consistent evaluation framework across all hourly initialization cycles. 

The HRRR’s fine spatial and temporal resolution makes it a critical tool for forecasters \citep{Dowell2022}. This reliance has driven significant development and improvement of the HRRR over the years. The LSTM-ViTs introduced herein are trained on three versions of the HRRR: HRRRv2 (January 2018 to 11 July 2018), HRRRv3 (12 July 2018 to 1 December 2020), and HRRRv4 (2 December 2020 to December 2023). A detailed list of the meteorological variables from the HRRR used as features in training the LSTM-ViTs is provided in Table \ref{tab:combined_features}. 

\subsection{Data Curation}

Detailed descriptions of mesonet instrumentation, quality control, station triangulation, geographic feature extraction, time encoding, and forecast error computation are provided in E25 and are not repeated here for brevity. The LSTM-ViT models are trained on data from the beginning of 2018 to the end of 2022 and validated on data from 2023. All models are then tested on data from 2024 and 2025 to capture seasonal and sub-seasonal model performance metrics.

\section{Model}

\begin{figure}
    \centering
    \includegraphics[width=36pc]{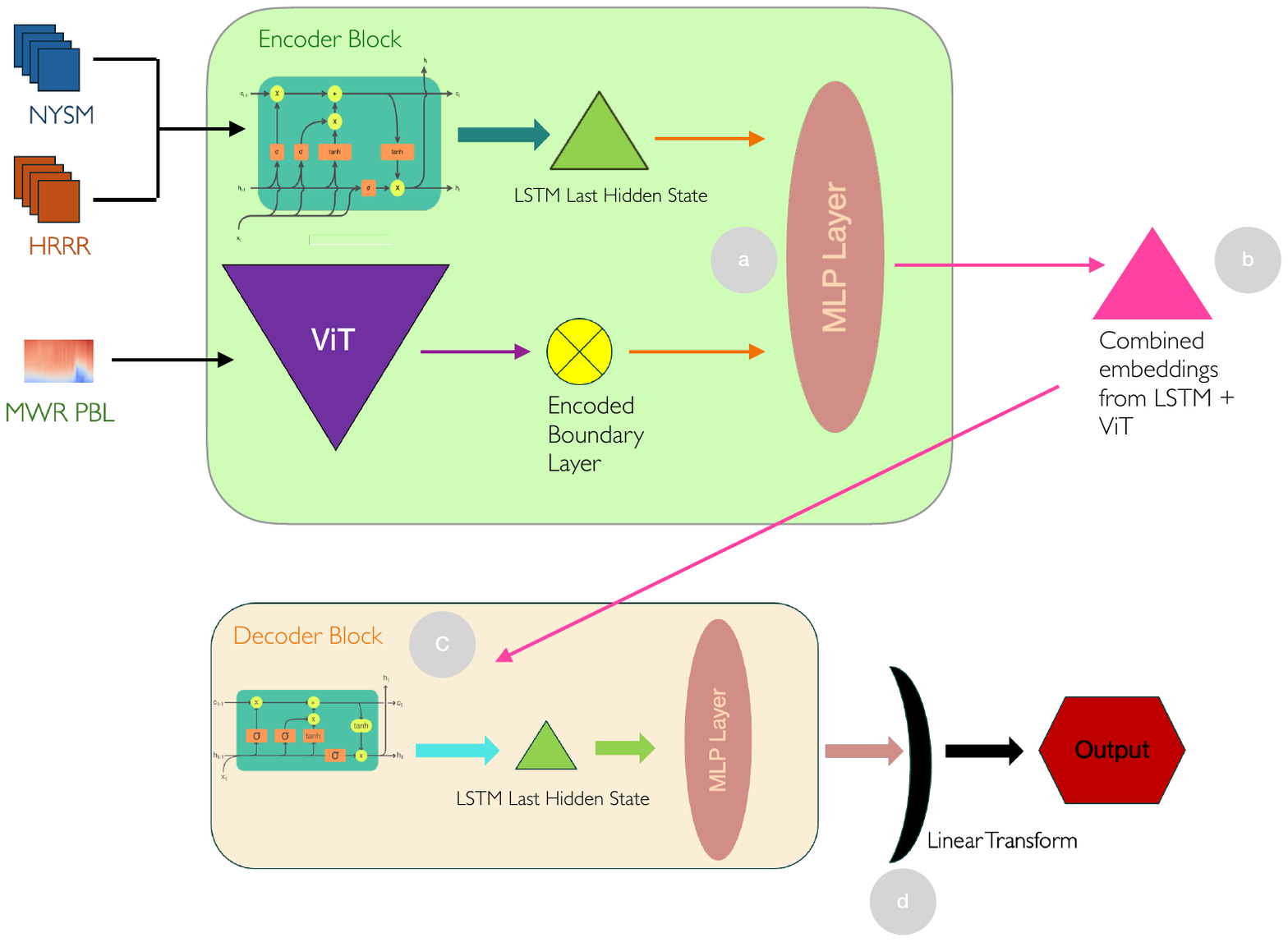}
    \caption{This graphic illustrates the LSTM+ViT encoder–decoder workflow at a high level.}
    \label{fig:entire-model}
\end{figure}

The LSTM-ViT architecture combines two complementary modeling paradigms that address distinct aspects of forecast-error prediction, and, although both architectures operate on temporally ordered data, they serve different purposes. The LSTM models the temporal evolution of surface observations and NWP forecasts, preserving the sequential and causal structure of forecast error prediction. In contrast, the ViT analyzes vertically resolved MWR profiles, using attention mechanisms to represent relationships among observations at different altitudes and times within the PBL. The model then combines the resulting latent representations into a unified description of the atmospheric state. Finally, an LSTM decoder uses this fused representation to generate forecast-error predictions across the forecast horizon. This design preserves the temporal forecasting capabilities of the original LSTM framework while incorporating information about the vertical structure and evolution of the lower atmosphere that surface observations alone cannot provide.

\subsection{LSTM Encoder}

In this study, we employ the LSTM encoder–decoder architecture introduced in our prior work (E25). We selected an LSTM architecture because forecast error evolves as a temporally dependent process influenced by the recent history of atmospheric observations, NWP forecasts, and model biases. The LSTM \citep{Sepp1997} extends the Recurrent Neural Network (RNN) by incorporating input, forget, and output gates that regulate information flow and mitigate vanishing gradients \citep{Sepp1997}, enabling the model to retain the long-range temporal structure critical for atmospheric prediction tasks. After processing the sequence of mesonet observations and HRRR forecast variables, the LSTM encoder compresses the recent atmospheric surface and forecast history into its final hidden state. This latent representation summarizes the temporal evolution of conditions relevant to forecast-error development and serves as input to the subsequent components of the hybrid framework, as seen in Fig~\ref{fig:entire-model}.

\subsection{Vision Transformer Encoder}

\begin{figure}
    \centering
    \includegraphics[width=33pc]{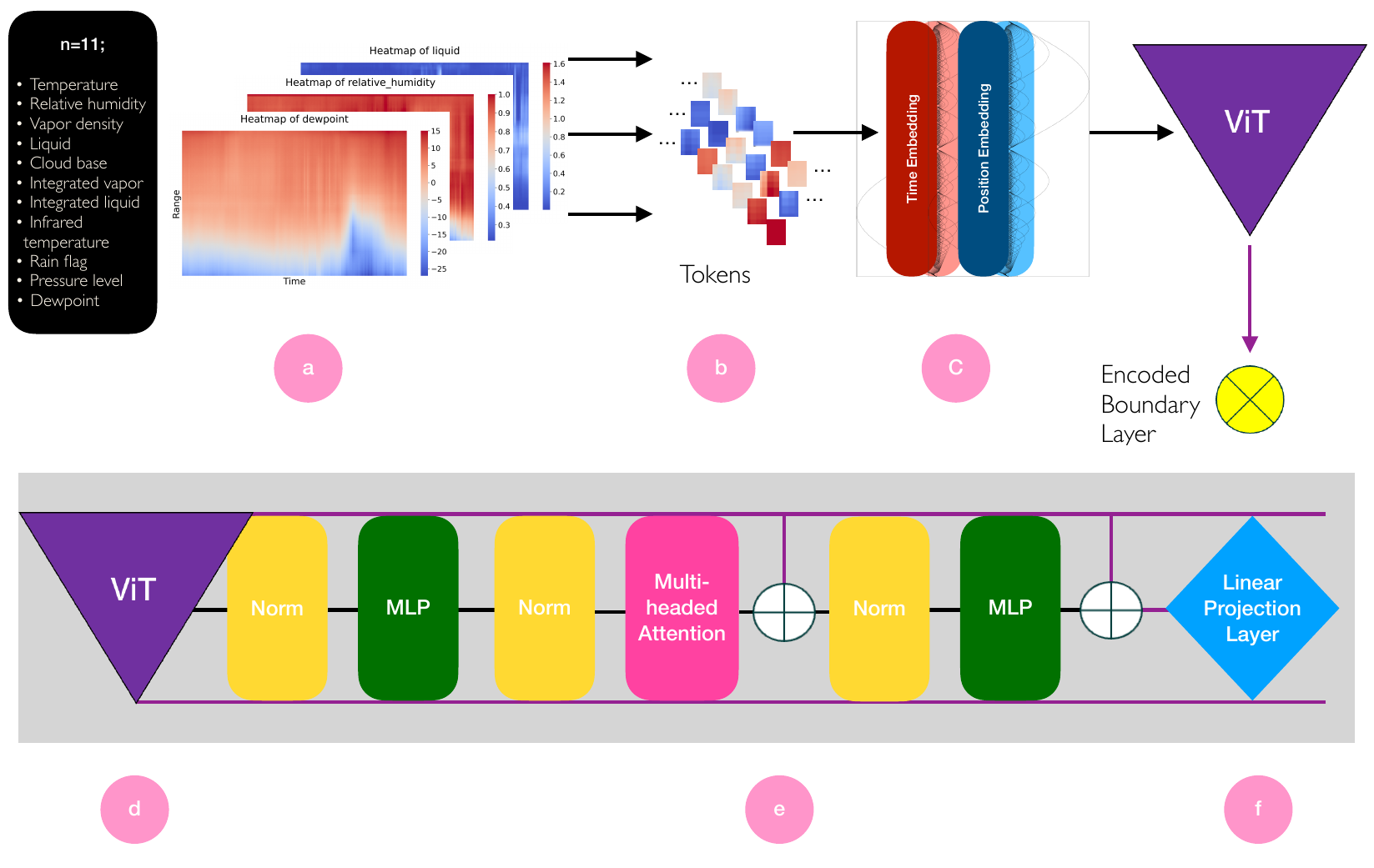}
    \caption{The structure of the Vision Transformer (ViT) encoder unit as implemented. Panel (a) illustrates the set of microwave-radiometer variables used as input, while panel (b) shows the construction of tokens from the time–height observation tensor. Panel (c) depicts the addition of learned temporal and positional embeddings, which provide information about the observation time and vertical location of each token. The ViT encoder aggregates information across all times and heights through self-attention, producing a latent representation of the recent atmospheric boundary-layer state. The gray block (panels d–f) illustrates the internal structure of the ViT encoder, following \citet{dosovitskiy2020}. The operations ``$+$'' denote element-wise addition associated with residual connections. The final linear projection layer maps the encoded representation into a fixed-length context vector, referred to herein as the encoded boundary layer, which is subsequently fused with the LSTM representation for forecast-error prediction.}\label{fig:vit-encoder}
\end{figure}

The widespread success of the transformer architecture introduced by \citet{vaswani2017} led to the development of ViT models for computer vision by \citet{dosovitskiy2020}.  ViTs model relationships between regions of an image as a sequence using attention mechanisms.  Unlike RNN networks, the attention mechanisms within transformers process sequence elements in parallel rather than sequentially.  As discussed by \citet{vaswani2017, dosovitskiy2020}, multi-headed attention allows the model to simultaneously focus on multiple relationships or patterns within the data. This allows the ViT to encode how each pixel (observed atmospheric condition) relates to each other regardless of space and time, more accurately encoding the most relevant information of the vertical profile to be fed to the LSTM for forecast error prediction.

\subsubsection{Input Representation}

The previous hour of MWR observations forms a three-dimensional tensor with dimensions Time $\times$ Height $\times$ Variable (Fig.~\ref{fig:vit-encoder}(a)). Because ViTs operate on sequences of tokens rather than multidimensional tensors, we flatten the Time $\times$ Height dimensions of the MWR tensor, $x \in \mathbb{R}^{T \times H \times V}$, into a sequence of $N=T\cdot H$ tokens, producing $x_p \in \mathbb{R}^{N \times V}$ (Fig.~\ref{fig:vit-encoder}(b)). Each token contains the complete set of MWR variables observed at a specific time and height level, enabling the self-attention mechanism to model relationships between observations occurring at different altitudes and times throughout the profile sequence. 

\subsubsection{Learned Position and Time Embeddings}

Because ViTs process inputs in parallel rather than sequentially, we encode both the position and time associated with each MWR observation by adding learned positional and temporal embeddings to each token (Fig.~\ref{fig:vit-encoder}(c)). Both embeddings are implemented as trainable parameters with dimensions $(1, N, HV)$, where $HV$ is the hidden dimension of the ViT. The positional embedding encodes the ordering of tokens within the sequence, while the temporal embedding provides information about absolute temporal context (e.g., hour of day or seasonal cycle). Together, these embeddings allow the model to represent the spatial and temporal dependencies that govern PBL evolution; without them, the model would have no explicit information about the temporal order or relative position of the MWR observations.

\subsubsection{Token Projection and Encoding}

The ViT encoder projects each token into the model embedding space using a normalized multilayer perceptron (MLP) projection (Fig.~\ref{fig:vit-encoder}(d)). The model then appends a learnable prediction token that, analogous to the class token of \citet{dosovitskiy2020}, aggregates information from the entire sequence and serves as a compact summary of the encoded atmospheric profile.

The resulting sequence passes through a standard ViT encoder consisting of alternating multi-head self-attention and MLP blocks, with layer normalization preceding each block and residual connections following each block (Fig.~\ref{fig:vit-encoder}(e); \citealp{dosovitskiy2020}).

\subsubsection{Linear Projection Layer}

The final novel component of the ViT encoder is the linear projection layer (Fig.~\ref{fig:vit-encoder}(f)). After processing the input sequence, the encoder extracts the final prediction token, which serves as a compact representation of the atmospheric profile. A learned linear projection then maps this representation into the hidden-state space of the LSTM decoder. Because the ViT produces a single embedding per sample whereas the decoder contains multiple LSTM layers, the framework replicates the projected context vector across the decoder-layer dimension so that each layer receives the same encoded representation of the MWR profile sequence. Thus, the projection layer functions as an interface between the ViT encoder and LSTM decoder, transforming the encoded MWR representation into a latent state compatible with the recurrent network.

\subsection{MLP Fusion}

After the LSTM and ViT encoders produce their latent representations, the model combines these complementary sources of information into a single hidden state. The LSTM encoder provides a temporally structured representation of surface observations and HRRR forecasts, while the ViT encoder provides an attention-based representation of the most recent hour of MWR profiler data. Because these encoders operate in different representational spaces, a fusion mechanism is required before their outputs can be used by the decoder.

To perform this integration, we concatenate the LSTM and ViT embeddings and pass the result through a ``fusion" MLP (Fig.~\ref{fig:entire-model}(a)). The MLP maps the combined representation into the LSTM decoder space, applies a LeakyReLU activation function to allow represention of nonlinear relationships between surface meteorology and vertical structure, and uses dropout to reduce overfitting and discourage over-reliance on either encoder.

The MLP produces a unified hidden state that integrates both surface and vertically resolved atmospheric information (Fig.~\ref{fig:entire-model}(b)). This fused representation is then passed to the decoder (Fig.~\ref{fig:entire-model}(c)), enabling forecast-error predictions that leverage both surface and vertical atmospheric structure.

\subsection{LSTM Decoder}

The remainder of the LSTM decoder follows the formulation detailed in E25. The decoder receives the ViT-derived context vector together with the encoded representation of mesonet observations and HRRR forecast history produced by the LSTM encoder. These representations initialize the decoder hidden state and provide the information required to generate future forecast-error predictions.

Although both the ViT and LSTM process temporally ordered data, they serve distinct purposes. The ViT extracts spatiotemporal features from the previous hour of vertically resolved MWR observations, while the LSTM decoder models the temporal evolution of forecast error across future lead times. This separation allows the ViT to focus on learning a representation from profile data while the decoder learns how forecast errors evolve through the forecast horizon.

The decoder generates predictions sequentially for each forecast hour, updating its hidden and cell states at each step. A fully connected output layer maps the resulting latent state to the forecast-error prediction for the corresponding forecast hour \citep{bishop2023}. By recursively propagating information through its hidden and cell states, the decoder captures dependencies among forecast hours and produces a coherent sequence of forecast-error predictions.

\subsubsection*{Linear Post Processing Function}

Lastly, as in E25, we apply linear post-processing (black crescent, Fig.~\ref{fig:entire-model}(d)) to tailor the LSTM output to the individual NYSM station, forecast hour, and forecast error variable of interest. The coefficients used for the linear post-processing calculations are determined using the validation fold of the data and are stored in a look-up table for testing and inference use. 

\subsection{Training Stability}

As in E25, we used several standard ML techniques to manage training stability and control overfitting. We implemented early stopping (patience = 8) to terminate training when validation performance ceased improving, thereby reducing overfitting and limiting the accumulation of prediction errors across lead times. We also applied a ReduceLROnPlateau learning-rate scheduler (patience = 4), which automatically reduced the learning rate whenever the validation loss stopped improving for several consecutive epochs. This adaptive strategy allowed the optimizer to take smaller parameter updates as training approached convergence and improved training stability. Finally, we controlled model complexity through hyperparameter tuning; Table~\ref{tab:hyperparameters} summarizes the hyperparameters we used in the LSTM-ViT architecture.

\begin{table}[h!]
\centering
\begin{tabular}{ll}
\hline
\textbf{Hyperparameter} & \textbf{Value} \\
\hline
Batch Size              & 70 \\
Learning Rate           & $9 \times 10^{-6}$ \\
Number of Layers        & 3 \\
Hidden Units (LSTM)     & 1728 \\
Sequence Length         & 15 \\
Number of Heads         & 11 \\
Hidden Dimension (ViT)  & 7260 \\
MLP Units (Encoder)     & 1032 \\
Regularization          & $1 \times 10^{-15}$ \\
Attention Dropout       & $1 \times 10^{-12}$ \\
Optimizer               & AdamW \\
Scheduler               & ReduceLROnPlateau (factor=0.1, patience=4) \\
Early Stopping          & Patience = 8 epochs \\
MLP Units (Decoder)     & 1500 \\
$\alpha$ (Loss Function) & 2.0 \\
\hline
\end{tabular}
\caption{Hyperparameters for the Hybrid LSTM-ViT model used in this study.}
\label{tab:hyperparameters}
\end{table}

\section{Results}

This section evaluates whether vertically resolved MWR observations improve HRRR forecast-error prediction relative to the surface-based LSTM framework from E25. We compare the hybrid LSTM-ViT and baseline LSTM models across the 17 NYSM profiler-network stations for accumulated precipitation, 10-m wind speed, and 2-m temperature errors. We focus on three questions. First, does the LSTM-ViT improve forecast-error prediction skill relative to the baseline LSTM? Second, do these improvements vary by forecast lead time, variable, season, time of day, or region? Third, are the largest gains consistent with forecast regimes in which vertical PBL structure should provide information beyond surface observations alone? Together, these comparisons quantify whether MWR profiles add predictive value to surface observations and HRRR forecast predictors, and they identify the conditions under which vertically resolved information most improves data-driven forecast-error prediction.

\subsection{Precipitation Error}

In E25, LSTM precipitation-error predictions degraded during warm-season convective events dominated by vertical motion and instability, while topography and storm frequency exerted secondary influences. The LSTM also exhibited asymmetric behavior: it accurately captured the magnitude of positive errors (wet bias) but underestimated negative magnitudes (dry bias), despite correctly identifying most negative-error cases. Additionally, the model consistently overpredicted small-magnitude precipitation errors. We developed the LSTM-ViT architecture to augment the LSTM's ability to represent vertical atmospheric structure and improve performance in convective and dynamically complex environments. The results presented herein evaluate the skill of this augmented framework. The following stations were omitted from this analysis due to erroneous model output: BUFF, BELL, ELLE, HFAL, MANH, TANN, and WARW.

\begin{figure*}
    \centering
    \includegraphics[width=33pc]{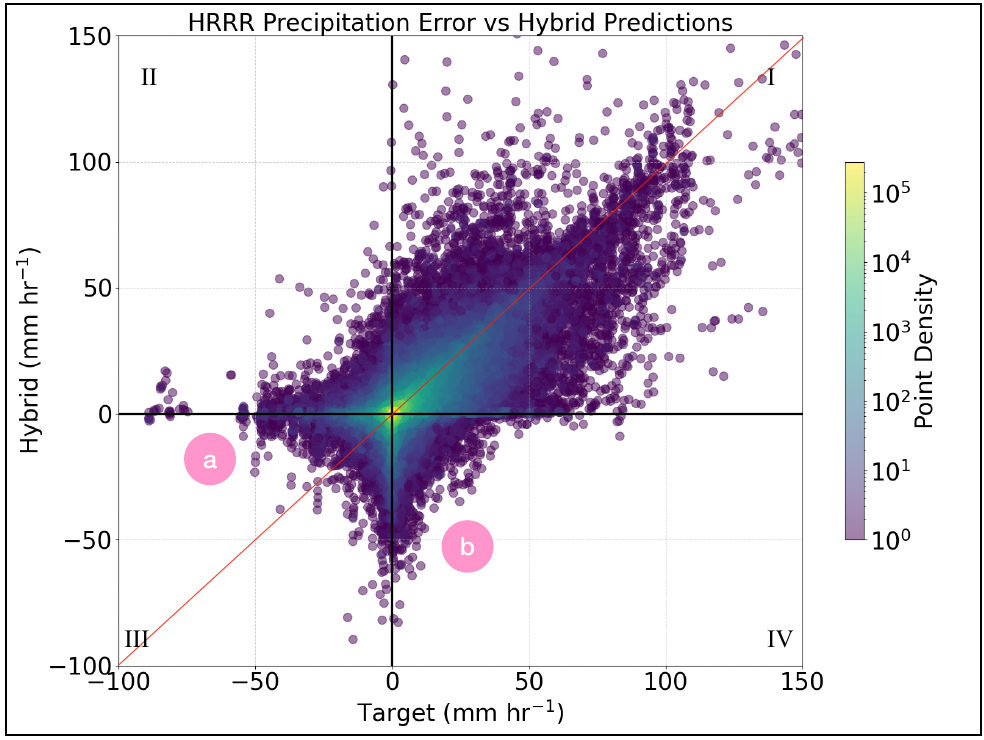}
    \caption{Scatterplot of the precipitation error across the NYSM network and all forecast hours, with the x-axis representing the true target error and the y-axis showing the corresponding Hybrid-predicted error. The red diagonal line indicates the 1:1 line, where perfect predictions would lie.}
    \label{fig:precip_scatter}
\end{figure*}

Figure~\ref{fig:precip_scatter} compares observed and predicted precipitation errors, with the red diagonal indicating the 1:1 line of perfect agreement. Within $\pm 5~\mathrm{mm~hr^{-1}}$, approximately 89\% of predictions lie on or near the 1:1 line, representing a 10\% improvement relative to the LSTM (E25). The LSTM-ViT model still exhibits asymmetric prediction skill, accurately capturing positive precipitation errors (wet biases; Fig.~\ref{fig:precip_scatter}(a), quadrant I) while underestimating the magnitude of negative errors (dry biases; quadrant III). However, this asymmetry is substantially weaker than in the LSTM framework (E25). The pronounced over-sensitivity to small positive errors present in the LSTM, which E25 primarily attributed to missed timing or spatial extent of convective events, is largely eliminated. The LSTM-ViT model retains some over-sensitivity to small negative errors (Fig.~\ref{fig:precip_scatter}(b)), typically associated with minor timing offsets in trace precipitation or synoptic-scale events, but this effect is also substantially reduced relative to the LSTM (E25).

\begin{figure*}
    \centering
    \includegraphics[width=28pc]{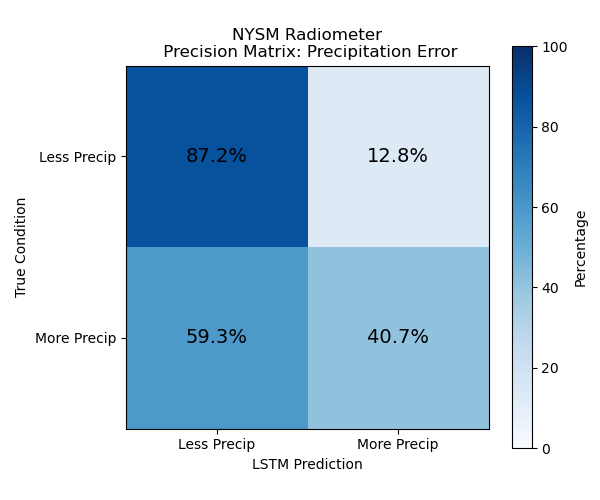}
    \caption{Confusion matrix summarizing the precision of Hybrid predictions for precipitation points across the entire NYSM and forecast hours. Rows indicate the true condition, and columns indicate the Hybrid’s prediction. More (less) precipitation translates to more (less) precipitation occurred than was forecast by the HRRR.}
    \label{fig:precip_confusion}
\end{figure*}

Figure~\ref{fig:precip_confusion} illustrates the LSTM-ViT model’s skill in detecting HRRR precipitation errors. The LSTM-ViT framework builds upon the LSTM’s ability to identify HRRR wet bias, improving precision by 4\% while maintaining an average wet-bias detection rate of 87\%. Dry-bias detection appears noisier than in the LSTM, which correctly identified dry bias in 75\% of cases but largely produced marginal negative-error predictions (E25). In contrast, the LSTM-ViT model generates more actionable negative-error forecasts, although timing errors and increased variability at small magnitudes  could be responsible for the apparent reduction in skill within the confusion matrix. Despite these limitations, the LSTM-ViT model correctly identifies HRRR dry bias in 41\% of cases.

\begin{figure*}
    \centering
    \includegraphics[width=33pc]{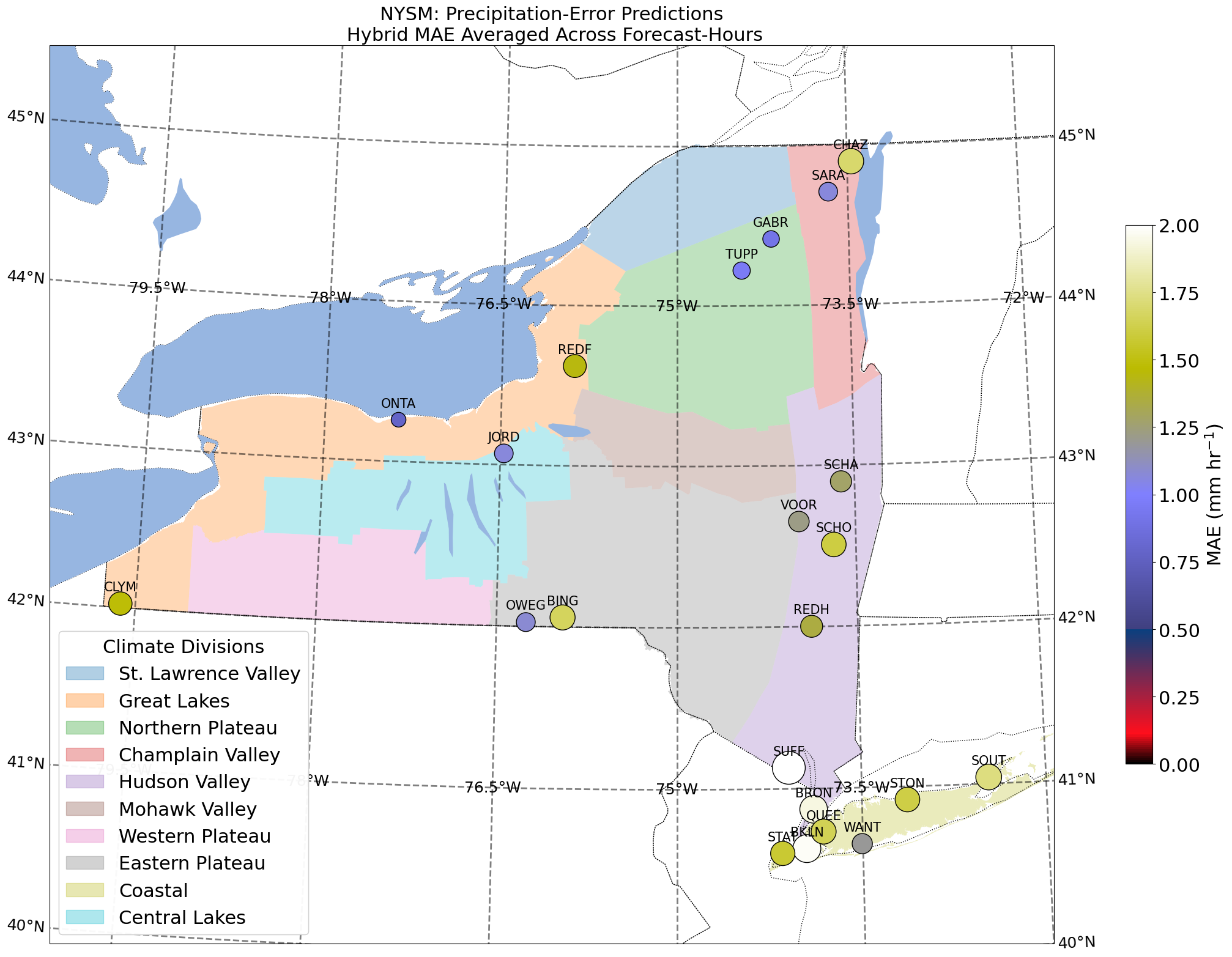}
    \caption{New York State MAE overlaid by NCEI climate division \citep{NCEI2015}. Each point represents the average Hybrid performance (MAE) for an NYSM station, averaged over all forecast lead times. The magnitude of the point is proportional to the MAE, where larger points translate to higher MAE.}
    \label{fig:precip_state}
\end{figure*}

Figure~\ref{fig:precip_state} shows the mean absolute error (MAE) of LSTM-ViT-model precipitation predictions across the NYSM Profiler Network. Spatial variability is relatively small, with most stations exhibiting MAE values between 0.75 and 2~mm~hr$^{-1}$. A regional maximum occurs within the Coastal climate division, including SUFF station, consistent with E25, where MAE is approximately 1~mm~hr$^{-1}$ higher than across the remainder of the domain. These regions also receive the highest annual precipitation totals within the NYSM network \citep{bader2023methodology, campbell_steenburgh2017, swain}. Enhanced land–sea contrasts and urban amplification \citep{swain} likely contribute to the elevated MAE observed in Fig.~\ref{fig:precip_state}. Although the LSTM-ViT model more effectively captures both positive and negative precipitation errors than the LSTM, the processes driving precipitation extremes remain difficult to predict at longer lead times, even with vertically resolved PBL information. These results suggest a persistent limitation in forecasting extreme precipitation associated with urban amplification, land–sea contrasts, and orographic effects characteristic of this region \citep{campbell_steenburgh2017, swain}.

\begin{figure*}
    \centering
    \includegraphics[width=33pc]{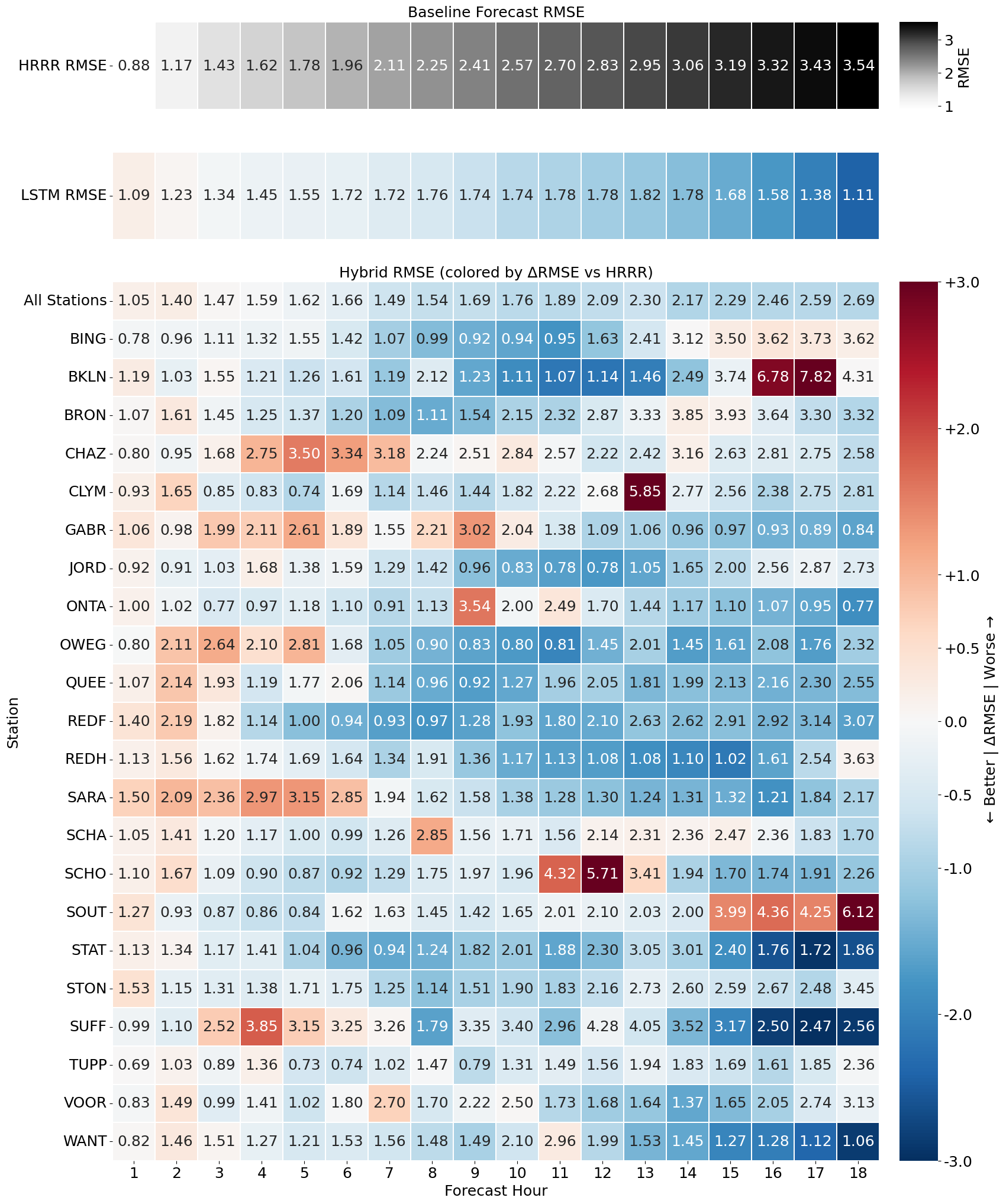}
    \caption{From top to bottom, panels show aggregate RMSE in $\mathrm{mm\,hr^{-1}}$ for the HRRR forecast, LSTM predictions, Hybrid model predictions, and then $\Delta$ RMSE comparison for each individual NYSM station in the Profiler network. In the HRRR panel, RMSE magnitude is represented by grayscale shading. In the subsequent ML panels, colors denote RMSE differences relative to the HRRR forecast, with red shading indicating higher RMSE and blue shading indicating lower RMSE than HRRR.}
    \label{fig:precip_wb}
\end{figure*}

Figure~\ref{fig:precip_wb} illustrates the relative improvement of LSTM-ViT-model predictions compared to the HRRR as a function of forecast lead time. The color shading is inspired by \citet{WeatherBench2}. Although the LSTM-ViT predicts forecast error rather than the meteorological variable itself, both the HRRR forecast and ML-corrected forecast are evaluated against the same observations. Consequently, the color scale represents relative improvement or degradation in forecast skill with respect to the baseline HRRR forecast. The top row shows the monotonic increase in HRRR RMSE with forecast lead time, a well-documented characteristic of NWP systems \citep{James2022}. The second row presents LSTM RMSE, highlighting its ability to systematically correct HRRR bias, as discussed in E25. The third row presents LSTM-ViT-model performance, beginning with an all-stations aggregate that enables direct comparison among the HRRR, LSTM, and LSTM-ViT frameworks.

The LSTM-ViT model achieves the lowest RMSE from forecast hours (FH) 1--12, outperforming both the HRRR and LSTM. Beyond FH 12, the LSTM achieves lower RMSE than both the HRRR and LSTM-ViT models, reflecting its strong ability to predict HRRR wet bias at longer lead times. This behavior also highlights a limitation of the LSTM-ViT model, which exhibits reduced consistency in predicting the sign and timing of errors. While the LSTM more reliably captures positive precipitation errors, the LSTM-ViT model provides greater utility for identifying high-impact and negative-error events, albeit with occasional temporal offsets that increase RMSE at longer lead times.

Overall, the LSTM-ViT models improve upon the HRRR in a manner that generally scales with forecast lead time. Several stations exhibit particularly large improvements at longer lead times, including STAT, SUFF, and WANT, all located within the Coastal and Lower Hudson Valley climate divisions where the analyzed precipitation extreme occurs. STAT and WANT display similar signatures, with substantial improvements at both short and long lead times, whereas SUFF exhibits more modest gains early in the forecast followed by a pronounced increase in skill after FH 15.

Most stations achieve their largest improvements at intermediate lead times (FH 4--12), likely reflecting the increased influence of PBL processes captured by the LSTM-ViT architecture. This behavior occurs across most of the network, except within the Northern Plateau and Champlain Valley climate divisions, which exhibit distinct error signatures. Stations in these regions show more modest improvements and occasionally perform comparably to or slightly worse than the HRRR at select lead times (e.g., CHAZ at FH 14; GABR at FH 9 and 10). Because RMSE at these sites is not systematically higher than elsewhere in the network, the results suggest that the HRRR already performs reasonably well in these regions or that residual errors arise primarily from timing and magnitude mismatches rather than persistent model bias.

\begin{figure*}
    \centering
    \includegraphics[width=36pc]{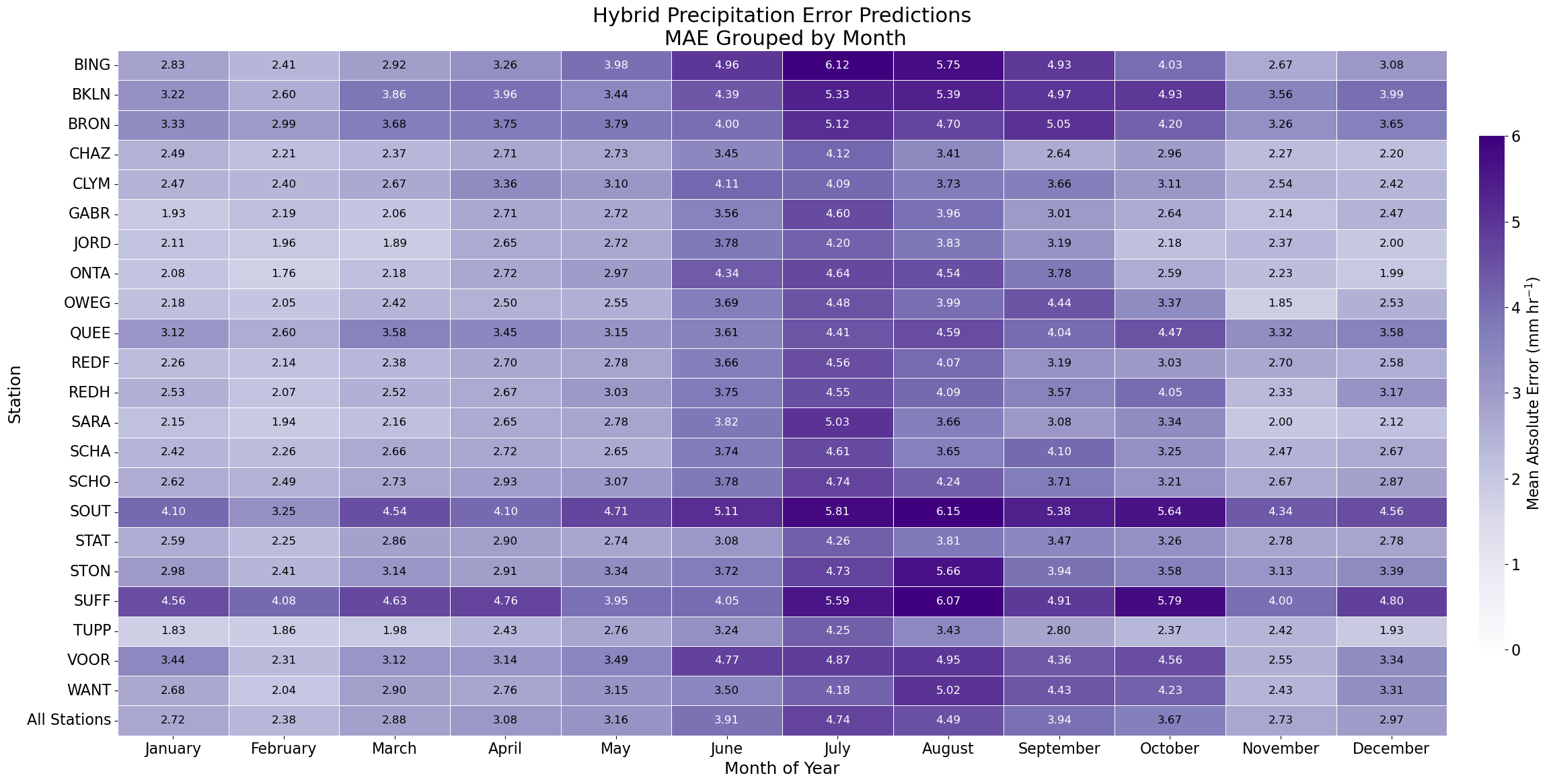}
    \caption{NYSM, MAE of LSTM-ViT precipitation-error predictions in $\mathrm{mm\,hr^{-1}}$, grouped by month. Panels are filtered to exclude zero-error LSTM-ViT predictions to better highlight model failure modes.}
    \label{fig:precip_year}
\end{figure*}

Figure~\ref{fig:precip_year} presents monthly MAE values ($\mathrm{mm~hr^{-1}}$) for LSTM-ViT precipitation-error predictions, filtered to exclude zero-error LSTM-ViT predictions to better highlight model failure modes. Many of the seasonal signatures identified in E25 persist in the LSTM-ViT architecture, although MAE is reduced by roughly half in most cases. Error exhibits strong seasonality, with the largest values occurring during summer (July–August), when MAE exceeds twice the yearly minimum (approximately $3~\mathrm{mm~hr^{-1}}$) and reaches a maximum of $6.15~\mathrm{mm~hr^{-1}}$ at SOUT in August. Stations within the Hudson Valley, Eastern Plateau, and Coastal climate divisions also exhibit coherent secondary maxima during winter (December–February), where MAE increases by approximately $1~\mathrm{mm~hr^{-1}}$ relative to annual minima. These regions correspond closely to the elevated MAE patterns shown in Fig.~\ref{fig:precip_state}.

Overall, incorporating vertically resolved PBL information substantially improves precipitation forecast-error prediction across a wide range of spatial and temporal scales. Relative to the baseline LSTM explored in E25, the LSTM-ViT architecture reduces systematic bias, improves representation of high-impact and negative-error events, and produces more spatially coherent skill across much of the NYSM network. Although challenges remain in regions dominated by complex convective, coastal, and orographic processes; the results demonstrate that vertical atmospheric structure provides meaningful predictive information for precipitation forecast-error growth and motivate further exploration of vertically informed ML architectures for environments characterized by strong instability, mesoscale forcing, and rapidly evolving PBL dynamics.

\subsection{Wind Error}

In E25, LSTM wind-error predictions exhibited a broadly consistent diurnal signature across geographic regions, reflecting atmospheric processes associated with PBL depth, vertical mixing, and orographic influences. Error was lowest prior to morning PBL spin-up and following afternoon mix-out, with enhanced improvement relative to the HRRR during the early evening hours after sunset. This pattern likely reflects increased LSTM skill under stable and well-mixed PBL conditions. Among the three predictors examined in E25, wind-error predictions exhibited the strongest covariance with the target and the most consistent performance across both over- and under-prediction regimes.

The LSTM-ViT model was developed to augment the LSTM’s representation of vertical atmospheric processes influencing near-surface wind forecasts, including PBL stratification, temperature inversions, vertical momentum transport, and convective circulations. By incorporating vertically resolved atmospheric structure, the LSTM-ViT framework aims to better capture wind forecast error associated with evolving PBL dynamics and terrain-modulated flow interactions. The following stations were omitted from this analysis due to erroneous model output: BKLN, CHAZ, ELLE, and QUEE.

\begin{figure*}
    \centering
    \includegraphics[width=36pc]{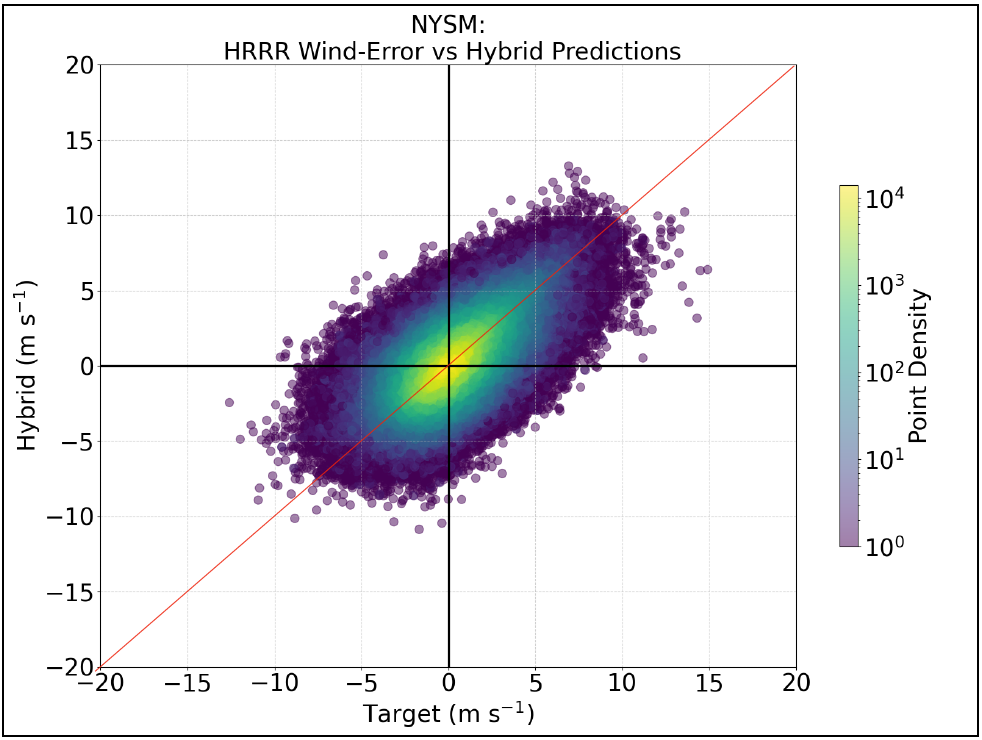}
    \caption{Scatterplot of the wind error across the NYSM network and all forecast hours, with the x-axis representing the true target error and the y-axis showing the corresponding LSTM-ViT-predicted error. The red diagonal line indicates the 1:1 line, where perfect predictions would lie.}
    \label{fig:wind_scatter}
\end{figure*}

Figure~\ref{fig:wind_scatter} compares observed and predicted wind errors, with the red diagonal denoting the 1:1 line of perfect agreement. Within $\pm~2~\mathrm{m~s^{-1}}$, approximately 92\% of predictions fall on or near the 1:1 line, indicating agreement comparable to that of the LSTM framework presented in E25. Consistent with E25, the LSTM-ViT model exhibits the strongest covariance with the target observations among the three forecast domains while maintaining relatively symmetric performance across both positive and negative error regimes.

\begin{figure*}
    \centering
    \includegraphics[width=33pc]{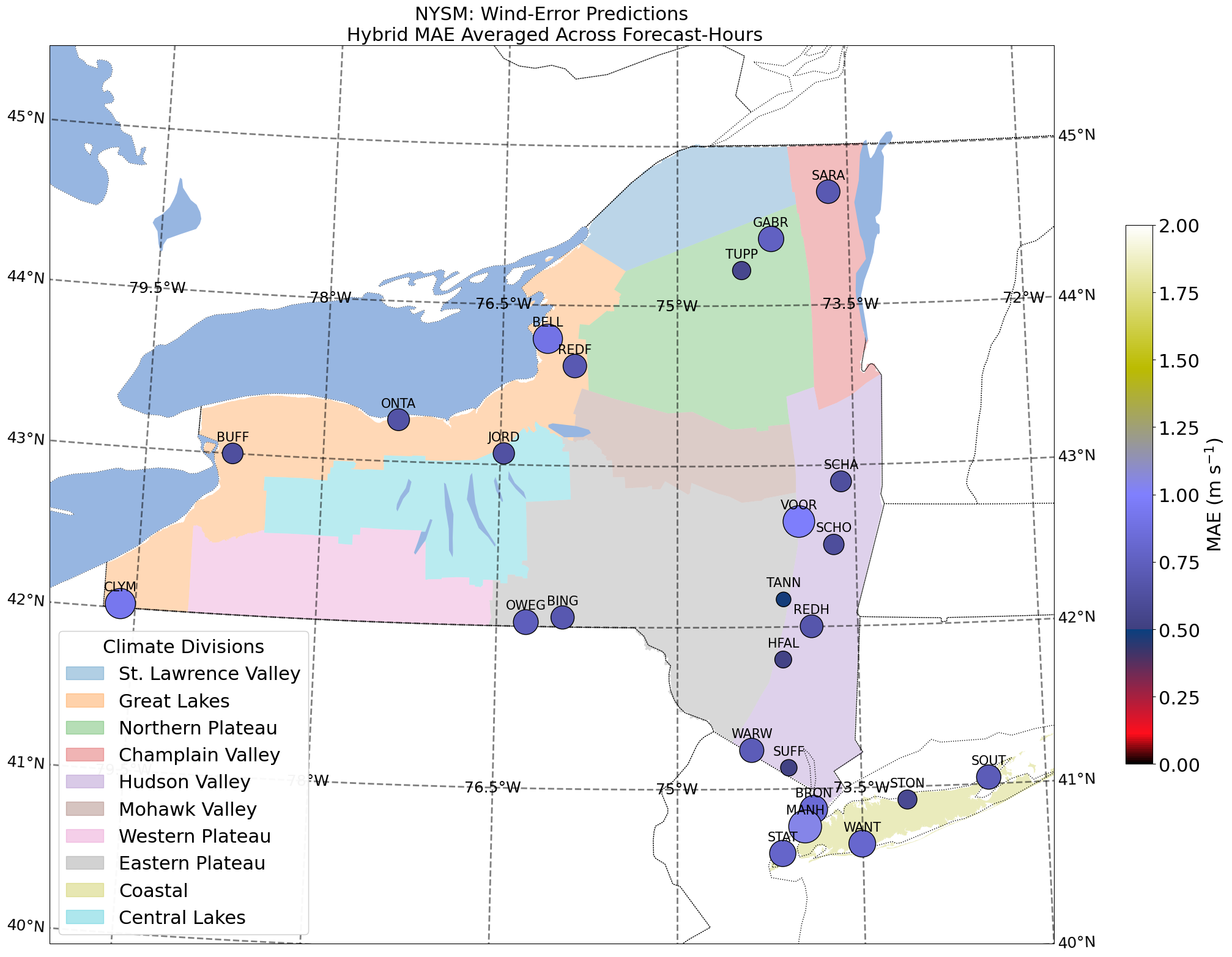}
    \caption{New York State MAE overlaid by NCEI climate division \citep{NCEI2015}. Each point represents the average LSTM-ViT performance (MAE) for an NYSM station, averaged over all forecast lead times. The magnitude of the point is proportional to the MAE, where larger points translate to higher MAE.}
    \label{fig:wind_state}
\end{figure*}

Figure~\ref{fig:wind_state} displays the MAE of LSTM-ViT-model wind predictions across the NYSM Profiler Network. Spatial variability is modest, with most stations exhibiting errors between 0.25 and $1~\mathrm{m~s^{-1}}$. No coherent geographic or regional error structure is evident, suggesting relatively uniform performance across the domain. This MAE distribution is broadly consistent with the station-level patterns reported for the LSTM framework in E25.

\begin{figure*}
    \centering
    \includegraphics[width=36pc]{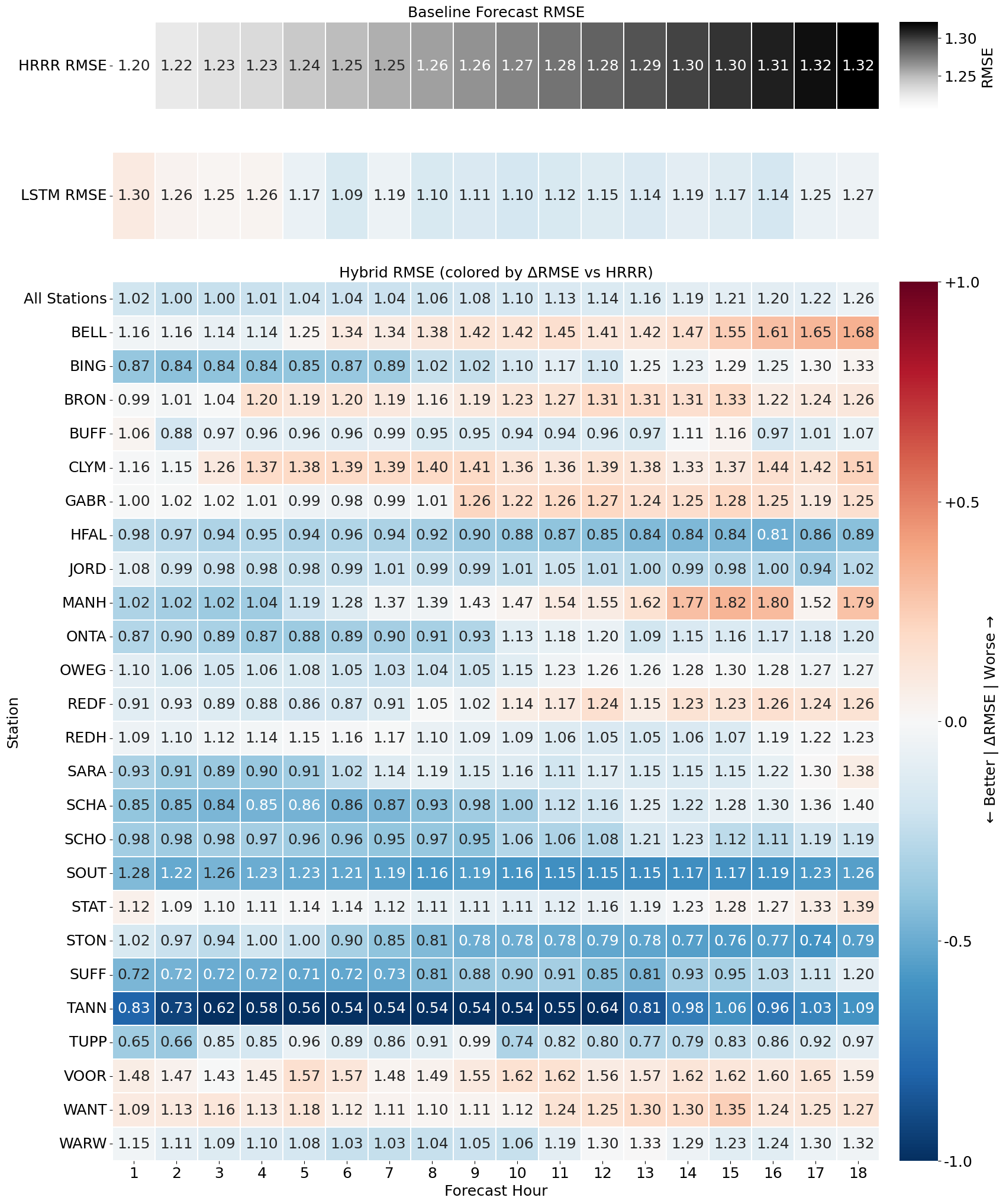}
    \caption{From top to bottom, panels show aggregate RMSE in $\mathrm{m~s^{-1}}$ for the HRRR forecast, LSTM predictions, LSTM-ViT model predictions, and then $\Delta$ RMSE comparison for each individual NYSM station in the Profiler network. In the HRRR panel, RMSE magnitude is represented by grayscale shading. In the subsequent ML panels, colors denote RMSE differences relative to the HRRR forecast, with red shading indicating higher RMSE and blue shading indicating lower RMSE than HRRR.}
    \label{fig:wind_wb}
\end{figure*}

As in Fig.~\ref{fig:precip_wb}, Fig.~\ref{fig:wind_wb} illustrates the relative improvement of LSTM-ViT-model predictions compared to the HRRR as a function of forecast lead time. The HRRR exhibits the expected monotonic increase in RMSE with lead time. The LSTM-ViT model maintains the lowest RMSE across all forecast horizons, with the largest improvements occurring at earlier lead times. In contrast, the LSTM performs worse than the HRRR at FH 1--4, improves upon the HRRR from FH 8--16 with skill comparable to the LSTM-ViT model, and becomes comparable to the HRRR at the final two forecast hours.

These results reinforce findings from E25, where the LSTM more effectively corrected systematic HRRR biases at longer lead times but struggled when forecast errors were dominated by rapidly evolving atmospheric processes. In contrast, the LSTM-ViT model consistently outperforms both the HRRR and LSTM, with the largest gains occurring during FH 1--10, similar to the precipitation-error domain. This enhanced short-range performance likely reflects the incorporation of vertically resolved PBL information, which improves representation of forecast-error growth associated with complex mesoscale and PBL processes.

Figure~\ref{fig:wind_wb} shows that greater inter-station variability emerges than in the precipitation-error domain. This behavior likely reflects the already strong baseline performance of the HRRR for near-surface wind magnitude, leaving less systematic error for the ML models to correct. In addition, terrain channeling, coastal interactions, and PBL stability strongly influence surface winds across NYS and may not be resolved consistently across all stations and atmospheric regimes. Consequently, forecast-error evolution varies substantially between locations, producing greater station-to-station variability in LSTM-ViT-model skill.

Nevertheless, many stations demonstrate notable improvements relative to the HRRR, particularly at shorter lead times, suggesting that vertical PBL structure provides meaningful predictive information for localized wind forecast-error growth. With the exception of BRON, CLYM, VOOR, and WANT, all stations exhibit at least some forecast horizons with improved performance relative to the HRRR. HFAL, JORD, ONTA, SOUT, STON, and especially TANN show particularly large RMSE reductions, whereas BUFF, REDH, STAT, and WARW exhibit more modest improvements. BELL, GABR, and REDF improve upon the HRRR at shorter lead times, but this advantage diminishes beyond approximately FH 5, 9, and 10, respectively.

\begin{figure*}
    \centering
    \includegraphics[width=36pc]{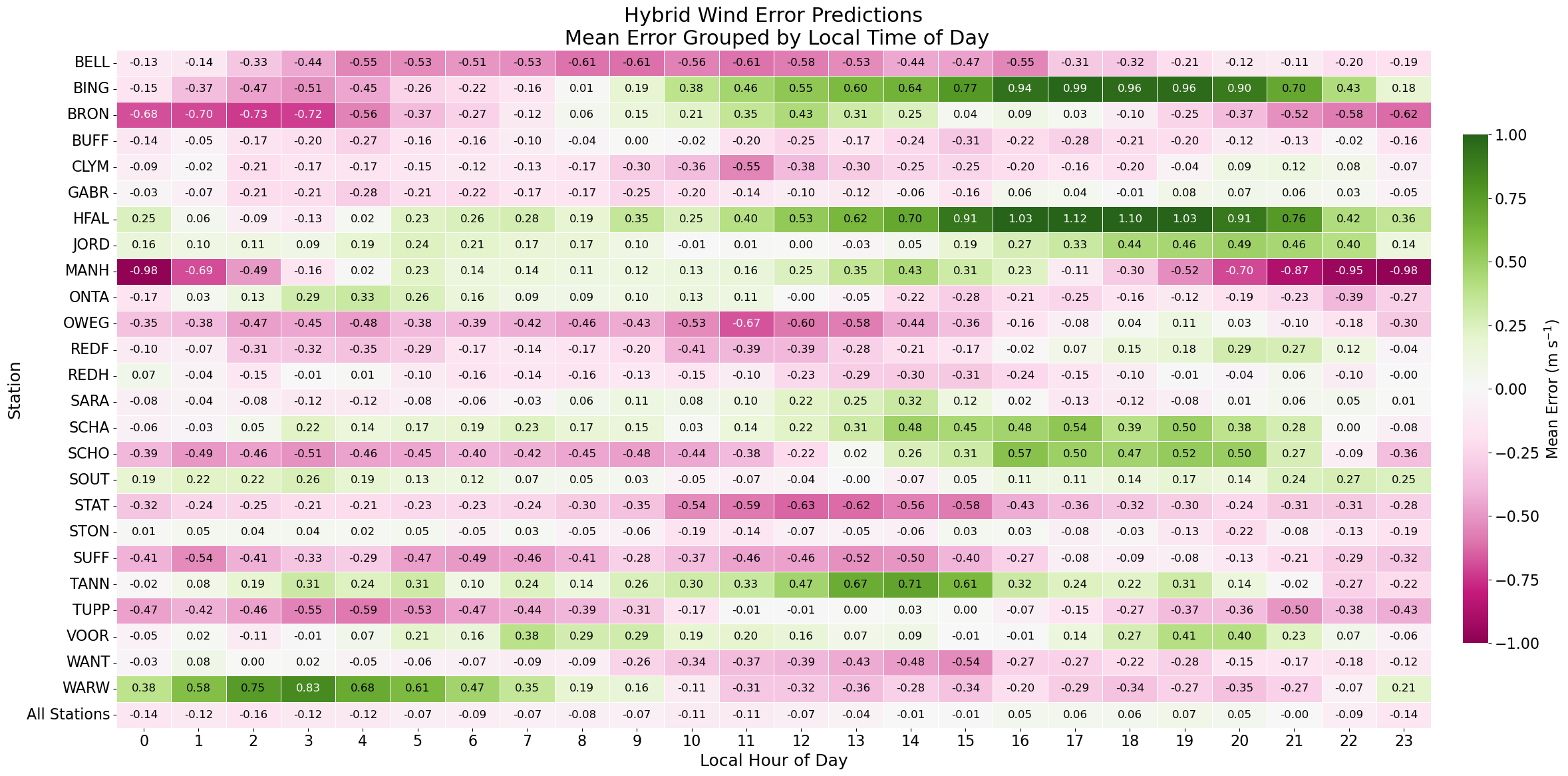}
    \caption{NYSM, MAE of LSTM-ViT wind-error predictions in $\mathrm{m~s^{-1}}$, grouped by local time of day. Panels are filtered to exclude zero-error LSTM-ViT predictions to better highlight model failure modes.}
    \label{fig:wind_day}
\end{figure*}

Figure~\ref{fig:wind_day} presents mean error ($\mathrm{m~s^{-1}}$) values grouped by local time of day for LSTM-ViT wind-error predictions, filtered to exclude zero-error LSTM-ViT predictions to better highlight model failure modes. The aggregate diurnal signal (bottom row) is broadly consistent with the temporal performance patterns identified in E25, particularly with respect to periods of relative skill degradation; however, both the magnitude and direction of the bias differ substantially between the LSTM and LSTM-ViT frameworks. Whereas the LSTM exhibited a persistent positive bias across most climate divisions, the LSTM-ViT model shows comparatively stable performance throughout the diurnal cycle. A weak negative bias occurs during the overnight and early morning hours (2200--0600), a slight positive bias emerges during the late afternoon (1600--1900), and a modest negative bias appears near solar noon. Overall, these results suggest that incorporating vertical PBL structure reduces the systematic overprediction observed in the LSTM baseline and produces a more temporally balanced representation of wind forecast error.

Unlike the results in E25, most stations exhibit subtle to pronounced negative bias. Several stations, including BING, HFAL, JORD, SCHA, and SCHO, display positive bias during the late afternoon and early evening, suggesting a tendency to overpredict wind forecast error during periods of enhanced daytime mixing and PBL transition. TANN and WARW also exhibit pronounced positive bias, although their temporal evolution differs from the broader network. Conversely, BELL, OWEG, STAT, SUFF, and WANT show substantially stronger underprediction, with the largest negative biases generally occurring near solar noon and during the pre-sunrise hours. A subset of stations, including HFAL, JORD, MANH, ONTA, SCHA, SOUT, TANN, and VOOR, further exhibit weak but consistent overprediction during the morning transition period (0500--1000). Collectively, these results highlight the highly station-dependent nature of wind forecast-error prediction and suggest that local PBL evolution, terrain interactions, and station-specific atmospheric regimes strongly influence LSTM-ViT-model behavior.

These results indicate that incorporating vertically resolved atmospheric structure substantially alters wind forecast-error prediction relative to the baseline LSTM framework. While the LSTM-ViT model demonstrates improved aggregate skill, particularly at shorter lead times, its error signatures become more localized, station-dependent, and temporally heterogeneous across the NYSM network. Unlike the broader regional structures evident in the LSTM framework, the LSTM-ViT architecture appears more sensitive to localized PBL evolution, terrain interactions, and mesoscale variability. This behavior suggests that vertical PBL information enables the model to resolve finer-scale processes influencing wind forecast-error growth, while also increasing sensitivity to highly localized atmospheric regimes. Overall, these findings highlight both the promise and complexity of vertically-informed ML frameworks for wind forecast-error prediction in geographically and dynamically complex environments.

\subsection{Temperature Error}

In E25, LSTM temperature-error predictions exhibited clear regional structures and distinct local-time signatures, likely reflecting atmospheric processes associated with PBL depth and vertical mixing. A subtle south--north gradient in model performance, with southern stations demonstrating greater predictive accuracy than northern stations, supports this interpretation. This pattern likely reflects systematic differences in PBL evolution between the maritime-influenced southern coast and the more continental, topographically complex northern regions. Northern New York also experiences more frequent and persistent snow cover, which can enhance surface-based temperature inversions and suppress PBL mixing. Because NWP systems often struggle to represent inversion strength, depth, and erosion, these conditions may introduce additional forecast-error variability that is more difficult for surface-based ML frameworks to predict.

To address these limitations, we developed the LSTM-ViT architecture to incorporate vertical profiles of temperature and latent variables, leveraging physical insights within an ML framework. The following results evaluate the predictive skill of this augmented architecture. The following stations were omitted from this analysis due to erroneous model output: GABR, HFAL, MANH, OWEG, SARA, SUFF, SCHA, SCHO, and TUPP.

\begin{figure*}
    \centering
    \includegraphics[width=36pc]{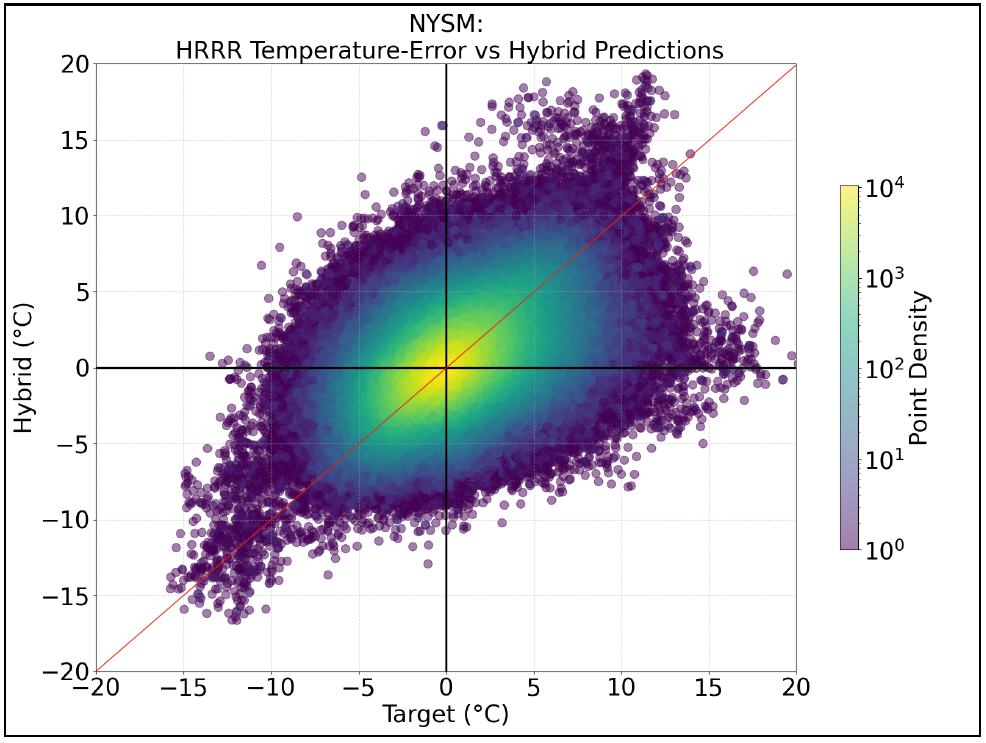}
    \caption{Scatterplot of the temperature error across the NYSM network and all forecast hours, with the x-axis representing the true target error and the y-axis showing the corresponding LSTM-ViT-predicted error. The red diagonal line indicates the 1:1 line, where perfect predictions would lie.}
    \label{fig:temp_scatter}
\end{figure*}

Figure~\ref{fig:temp_scatter} shows that within $\pm~2~\mathrm{^\circ C}$, approximately 73\% of predictions fall on or near the 1:1 line, representing a 1\% reduction in alignment relative to the LSTM (E25). A slight asymmetry remains evident in the outliers: negative errors align more closely with the 1:1 line, whereas positive errors tend to be over-predicted in magnitude. This behavior suggests that the LSTM-ViT model more effectively captures events in which observed temperatures exceed the HRRR forecast, while tending to underestimate the magnitude of events in which observed temperatures are colder than the HRRR forecast.

\begin{figure*}
    \centering
    \includegraphics[width=33pc]{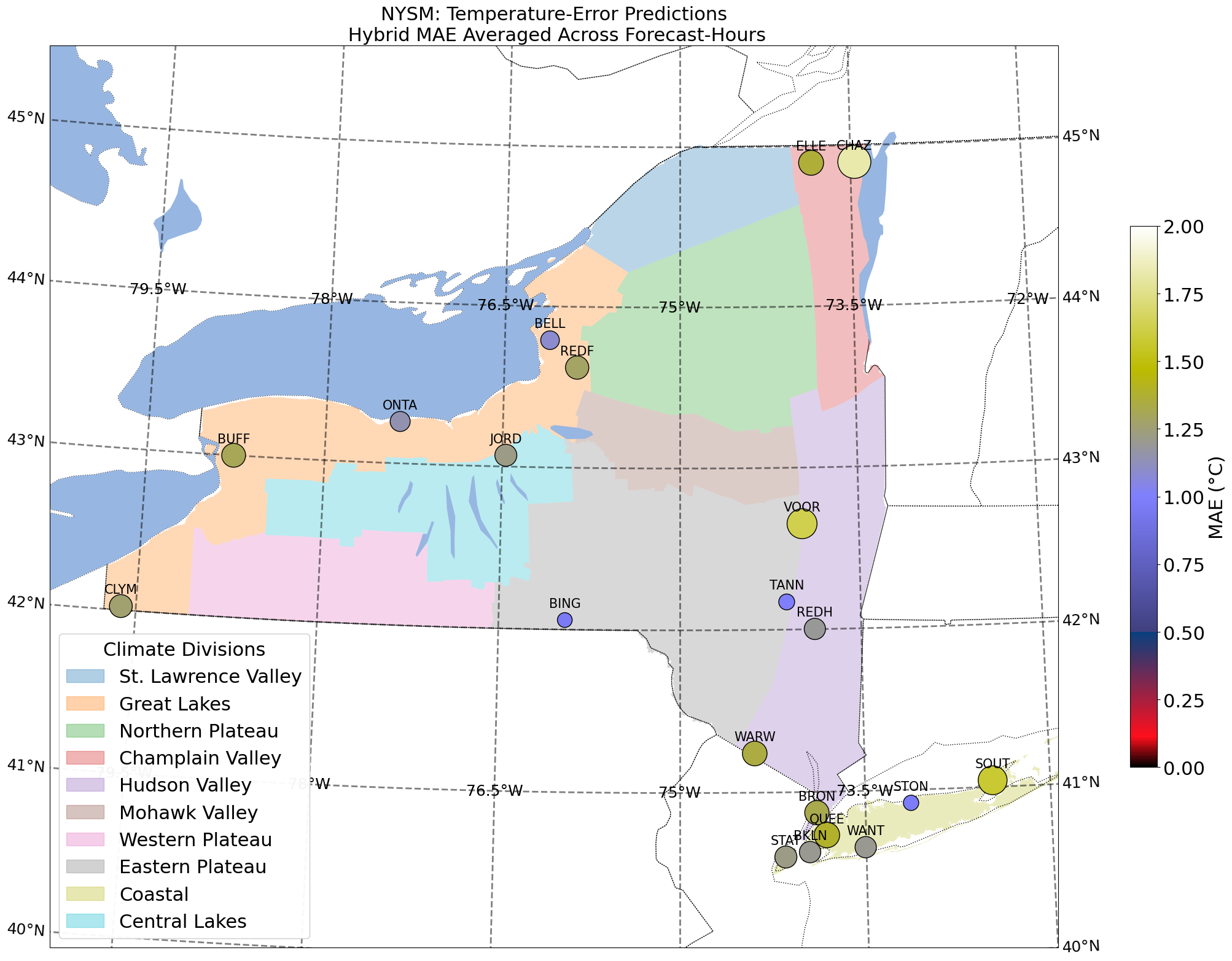}
    \caption{New York State MAE overlaid by NCEI climate division \citep{NCEI2015}. Each point represents the average LSTM-ViT performance (MAE) for an NYSM station, averaged over all forecast lead times. The magnitude of the point is proportional to the MAE, where larger points translate to higher MAE.}
    \label{fig:temp_state}
\end{figure*}

Figure~\ref{fig:temp_state} shows that spatial variability is modest across NYS, with most stations exhibiting errors between 0.75 and $2\mathrm{^\circ C}$. No coherent geographic or regional pattern emerges across the network. Excluding TANN, BING, and STON, which exhibit notably lower MAE, the remaining stations fall within approximately $1\mathrm{^\circ C}$ of one another, indicating relatively uniform performance across sites.

\begin{figure*}
    \centering
    \includegraphics[width=36pc]{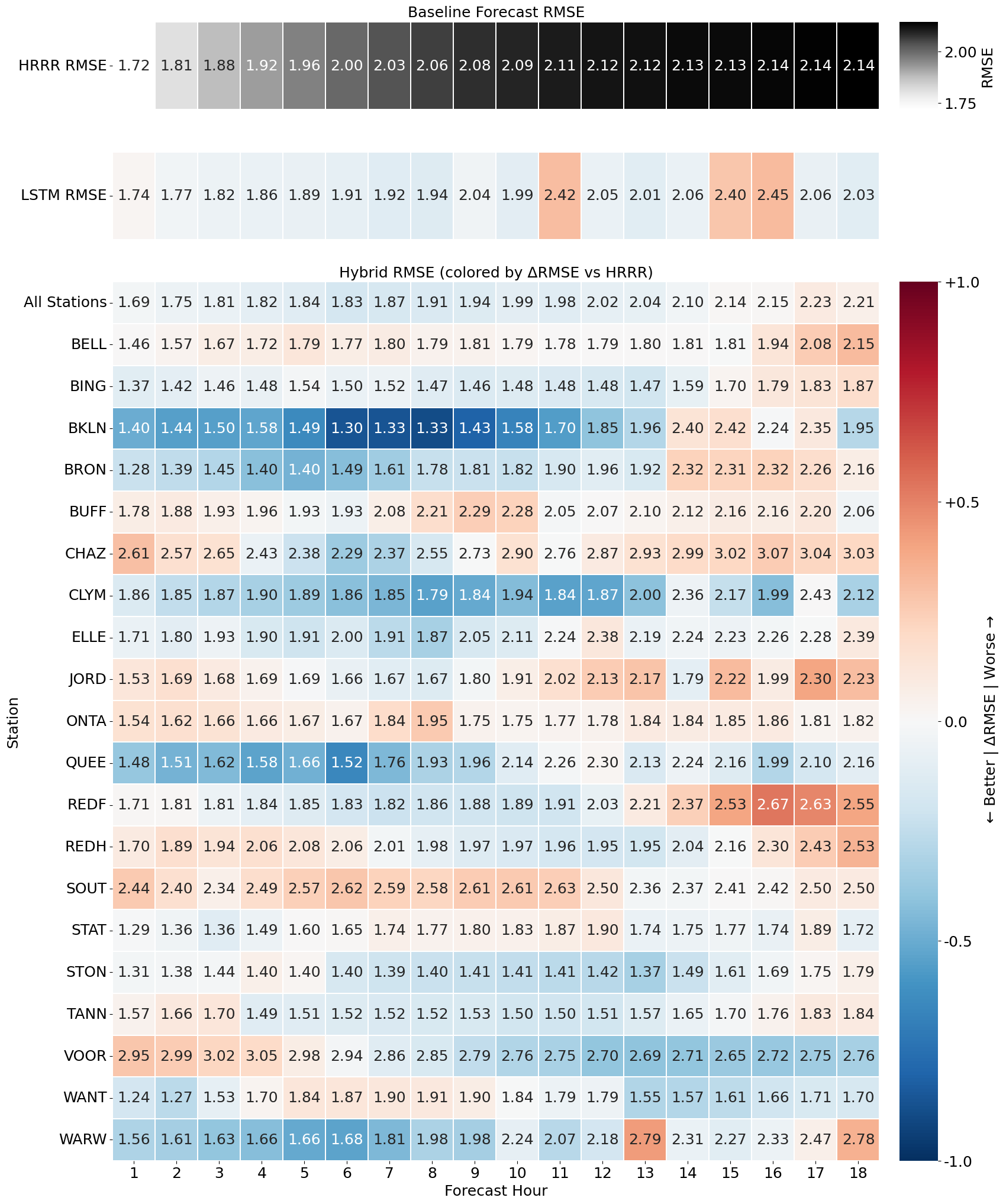}
    \caption{From top to bottom, panels show aggregate RMSE in $\mathrm{^\circ C}$ for the HRRR forecast, LSTM predictions, LSTM-ViT model predictions, and then $\Delta$ RMSE comparison for each individual NYSM station in the Profiler network. In the HRRR panel, RMSE magnitude is represented by grayscale shading. In the subsequent ML panels, colors denote RMSE differences relative to the HRRR forecast, with red shading indicating higher RMSE and blue shading indicating lower RMSE than HRRR.}
    \label{fig:temp_wb}
\end{figure*}

Comparing the top three rows in Fig.~\ref{fig:temp_wb}, the LSTM-ViT model exhibits slightly lower RMSE than the LSTM from FH 1--13. Beyond FH 13, the LSTM attains lower RMSE than the LSTM-ViT model and briefly underperforms the HRRR at FH 15--16. The LSTM-ViT model exhibits a near-monotonic increase in RMSE with lead time, whereas the LSTM displays greater variability, including a notable departure at FH 11 where it exceeds HRRR RMSE.

Consistent with E25, the LSTM generally provides skillful predictions, although its output tends to be smoother and occasionally temporally displaced relative to the target error. As with precipitation, the aggregate row in Fig.~\ref{fig:temp_wb} demonstrates that the LSTM-ViT model improves upon both the HRRR and LSTM primarily at early to intermediate lead times, with the largest gains occurring during FH 5--14. These improvements likely reflect the increased influence of PBL processes captured by the LSTM-ViT architecture. However, the magnitude of improvement is smaller than in the precipitation-error domain, highlighting the complexity of temperature-error prediction across NYS. 

Further analysis indicates that lead-time error signatures are highly station-specific rather than regionally coherent. BELL, BUFF, CHAZ, JORD, ONTA, REDH, SOUT, and WANT show minimal improvement relative to the HRRR across much of the forecast horizon, with some lead times exhibiting degraded performance. In most of these cases, RMSE is not substantially higher than elsewhere in the LSTM-ViT domain, suggesting that the HRRR already performs well at these locations or that increased temporal offsets and output variability contribute to the higher error. Several stations in the lower Hudson Valley and New York City metropolitan area (BKLN, BRON, QUEE, WARW) exhibit the largest RMSE reductions relative to the HRRR, particularly at shorter lead times (FH 1--9). These stations also performed well in the surface-based LSTM framework (E25), suggesting that local atmospheric processes governing forecast error are especially predictable at these locations. The LSTM-ViT framework further amplifies these improvements, likely because it can leverage vertical atmospheric information unavailable to the standalone LSTM.

\begin{figure*}
    \centering
    \includegraphics[width=36pc]{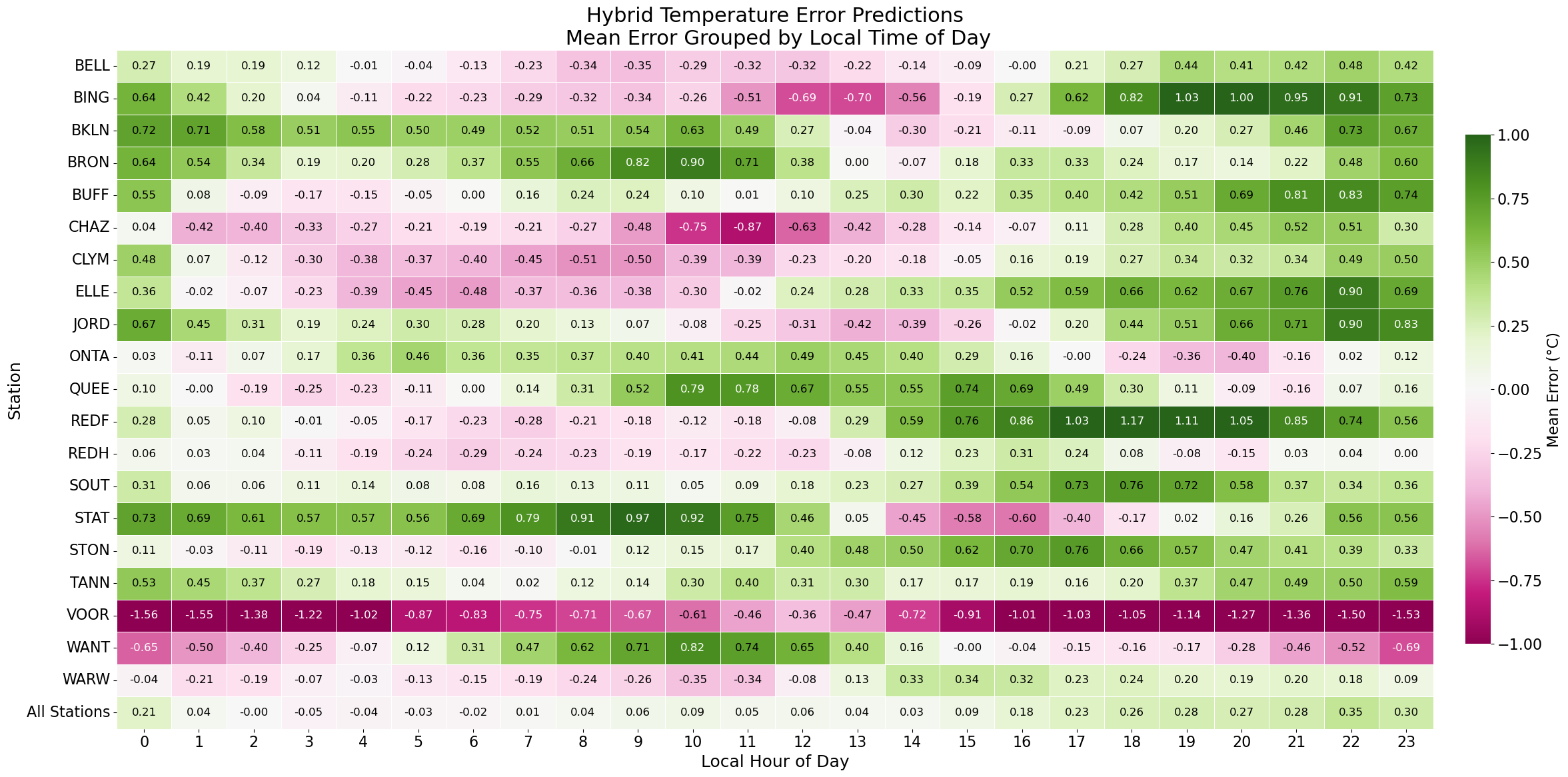}
    \caption{NYSM, MAE of LSTM-ViT temperature-error predictions in $\mathrm{^\circ C}$, grouped by local time of day. Panels are filtered to exclude zero-error LSTM-ViT predictions to better highlight model failure modes.}
    \label{fig:temp_day}
\end{figure*}

Referencing Fig.~\ref{fig:temp_day}, the aggregate diurnal signature (bottom row) differs from that identified in E25. The LSTM-ViT model transitions from slight under-prediction during the pre-sunrise hours (0200--0500) to modest over-prediction between 0700 and 1500 local time. From approximately 1600 through midnight, over-prediction gradually increases. Unlike the LSTM framework, the LSTM-ViT model performs comparatively well near solar noon, typically associated with peak PBL overturning, while skill degrades during the morning spin-up and evening mixing-out phases when turbulence and vertical structure are more dynamically complex.

Most NYSM stations broadly follow the aggregate signature, although many exhibit distinct local time-of-day patterns. A subset of stations (BKLN, BRON, JORD, ONTA, and STAT) displays a temporal structure that is largely anti-correlated with the aggregate behavior, transitioning from strong over-prediction overnight to modest under-prediction during the early afternoon. With the exception of JORD and ONTA, these stations are located within the NYC metropolitan area; however, the absence of a similar signature at other Coastal-division stations suggests that this behavior is not solely a regional effect.

SOUT, TANN, WANT, and VOOR exhibit particularly distinctive diurnal signatures. SOUT and TANN are characterized primarily by modest over-prediction throughout the day, although TANN displays greater temporal variability. In contrast, WANT and VOOR exhibit persistent under-prediction across much of the diurnal cycle, with VOOR maintaining the strongest and most consistent negative bias of any station in the network.

Overall, these results suggest that the LSTM-ViT architecture learns useful representations of vertically resolved thermodynamic structure from the MWR profiles. Incorporating this information improves temperature forecast-error prediction, particularly at shorter and intermediate lead times when PBL evolution strongly influences near-surface temperature variability. Although these gains are smaller than those observed for precipitation, the LSTM-ViT better captures localized and temporally evolving temperature-error signatures than either the HRRR or baseline LSTM. The station-specific nature of these errors highlights the complexity of temperature prediction across NYS, where terrain, urbanization, land--water contrasts, and PBL evolution interact across multiple scales. Collectively, these findings show that vertically informed ML architectures can better represent physically driven temperature forecast errors, while underscoring the continued challenge of resolving highly localized thermodynamic processes.

\section{Summary and Conclusions}

We trained independent hybrid LSTM-ViT models for each mesonet station and target variable using observations from the NYSM Profiler and Standard networks to predict HRRR hourly precipitation, 10-m wind speed, and 2-m temperature forecast errors. Models were trained on data from 2018--2023 and evaluated on data from 2024--2025. We employed an outlier-focused loss function to improve representation of rare, high-impact events and evaluated performance using RMSE, MAE, and mean error, with additional analyses by geography, time of day, season, and forecast lead time.

As in \citet{Evans2025}, forecast-error prediction skill exhibits coherent spatial and temporal patterns, with reduced performance often occurring during periods associated with complex mesoscale and PBL processes. Although these relationships do not establish causality, they are consistent with the interpretation that topography, convection, and PBL evolution contribute to forecast-error complexity. Relative to the standalone LSTM, the LSTM-ViT framework exhibits substantially less degradation in environments influenced by complex vertical atmospheric processes. The largest improvements occur at shorter forecast lead times, where the observed vertical atmospheric structure remains most representative of the future forecast environment. As lead time increases, the atmosphere evolves farther from the observed profiler state, reducing the predictive value of the MWR observations and diminishing the relative advantage of the LSTM-ViT architecture. Consistent with this behavior, station-level statistical analyses indicate that the LSTM-ViT framework provides a statistically significant improvement over the standalone LSTM at short- to-intermediate-range lead times, whereas differences become smaller and are generally not statistically significant at the longest lead times.

LSTM-ViT precipitation-error prediction consistently improves upon the LSTM baseline across most performance metrics. In particular, the LSTM-ViT model produces more realistic negative-error forecasts and exhibits reduced performance degradation during convective events. However, the framework retains an asymmetry in error prediction, with positive errors generally captured more accurately than negative errors. A second limitation is temporal displacement. Although the LSTM-ViT model often identifies the occurrence of forecast-error events, predictions are frequently shifted by one to two forecast hours. This misalignment inflates RMSE and MAE, particularly at longer lead times, while increasing output variability further contributes to performance degradation. Reducing temporal offsets and improving forecast consistency represent important directions for future work.

LSTM-ViT wind-error predictions demonstrate modest but consistent improvements over the baseline LSTM framework, particularly at shorter forecast lead times. The LSTM-ViT model retains some of the performance degradation associated with diurnal PBL evolution and transition periods; however, these signatures are more localized and exhibit different magnitudes and bias structures than those observed in the LSTM framework. This spatial heterogeneity likely reflects the inherent difficulty of predicting near-surface wind magnitude, where terrain effects, turbulent mixing, and mesoscale variability strongly influence forecast-error evolution. Although incorporating vertically resolved thermodynamic information improves predictive skill, wind forecast error appears less strongly constrained by the available MWR observations than the precipitation and temperature domains. Inclusion of vertically resolved wind profiles and more spatially consistent thermodynamic observations would likely further improve performance. Nevertheless, the LSTM-ViT framework consistently improves upon both the HRRR and baseline LSTM, particularly at shorter lead times, suggesting that it captures vertical atmospheric processes that contribute to near-surface wind forecast error.

LSTM-ViT temperature-error predictions exhibit more spatially and temporally coherent aggregate patterns than the LSTM baseline. Although overall performance between the two frameworks is broadly comparable, the LSTM-ViT model appears less sensitive to daytime PBL overturning and solar-radiation-driven variability, while exhibiting greater degradation under nocturnal conditions. This reduced nighttime skill likely reflects the increased complexity of near-surface thermodynamics after sunset, when local heat sources and sinks, stable boundary layers, and temperature inversions become more important. These processes remain difficult to resolve with NYSM profiler observations and may introduce biases that contribute to reduced model accuracy in nocturnal regimes.

These results suggest that the proposed LSTM-ViT framework is most effective at shorter forecast lead times, where it consistently outperforms the baseline LSTM in both the magnitude and directional accuracy of forecast-error prediction. This advantage is particularly evident in rapidly evolving atmospheric regimes, where vertically resolved PBL information improves representation of near-term forecast-error growth and variability.

At longer lead times, the LSTM framework explored in E25 often demonstrates more stable skill in correcting systematic HRRR biases, particularly for persistent and spatially coherent error structures. Consequently, the two architectures exhibit complementary strengths: the LSTM-ViT framework better resolves localized and dynamically evolving processes at shorter lead times, whereas the LSTM more effectively captures broader-scale and temporally persistent forecast biases at extended lead times. These findings suggest that vertically informed architectures and sequence-learning approaches may provide the greatest operational value when used as complementary components of a unified forecast-error prediction framework rather than as competing methodologies.

Although the LSTM-ViT framework advances beyond the LSTM baseline, important limitations remain. Future work should focus on developing a Bayesian Neural Network (BNN)-based Uncertainty Calibration and Reliability (UCR) framework for both architectures, enabling probabilistic characterization of forecast-error predictions and providing forecasters with actionable measures of confidence alongside deterministic guidance \citep{mansfield2025, clare2022explainable}.

Another limitation of the present study is the need for improved training stability across the modeling framework. Several stations were excluded from the analysis due to erroneous or unstable model output, indicating potential issues with data quality, observational representativeness, or model robustness. Profiler-data inconsistencies, calibration issues, and substantial missing data may reduce training effectiveness at some locations. In addition, many affected stations are not co-located with radiometers but instead rely on observations from sites within a 30-km radius, suggesting that limitations in observational coverage and instrument placement may restrict the model’s ability to accurately characterize local vertical atmospheric structure. These findings highlight the need for improved data quality control, enhanced observational coverage, and more robust training strategies in future work.

The LSTM–ViT framework provides station-specific, real-time guidance on the expected magnitude and direction of HRRR forecast error by jointly encoding surface observations and vertically resolved PBL structure. By combining temporal sequence learning from the LSTM with attention-based representations of MWR-derived atmospheric profiles from the ViT, the model captures forecast-error characteristics associated with PBL processes that are not fully represented in surface-based or purely sequential approaches. This capability enables more informed assessment of forecast uncertainty, particularly in environments associated with known model deficiencies and complex vertical variability.

The improved performance of the LSTM-ViT compared to the LSTM baseline demonstrates the value of combining vertical atmospheric context with temporal sequence learning for forecast-error prediction. The results highlight the potential of hybrid architectures to enhance forecast-error prediction in high-resolution NWP systems such as the HRRR, particularly at shorter lead times. This framework provides forecasters with a more physically informed assessment of forecast uncertainty at the point of use and can be readily extended to other mesonet–profiler networks and NWP systems.

\section*{Acknowledgments}
This material is based upon work supported by the U.S. National Science Foundation under Grant No. RISE-2019758. \\ This research is made possible by the New York State (NYS) Mesonet. Original funding for the NYS Mesonet (NYSM) buildup was provided by Federal Emergency Management Agency grant FEMA-4085-DR-NY. The continued operation and maintenance of the NYSM is supported by National Mesonet Program, University at Albany, Federal and private grants, and others. \\ The author thanks Dr. Andrew Fagg of the University of Oklahoma for his thoughtful reading of the manuscript and for providing valuable feedback that improved the final version of this work.

\section*{Data Statement}
HRRR data were accessed from the University of Utah MesoWest HRRR archive \citep{blaylock2017cloud} and from Amazon Web Services. \\ NYSM data can be requested through the New York State Mesonet website: \url{http://nysmesonet.org}. \\ The code used for data preprocessing, model training, and figure generation is publicly available at \url{https://github.com/shmaronshmevans/inference_ai2es_forecast_err}.

%%% BIBLIOGRAPHY %%%

\end{document}